\documentclass{article}
\usepackage{lineno,hyperref}
\usepackage[english]{babel}
\usepackage{graphicx}
\usepackage{booktabs} 
\usepackage{amsmath}
\usepackage{amssymb}
\usepackage{multirow}
\usepackage{subfigure}
\usepackage{eqnarray}
\usepackage{url}

\newcommand{\NO}[1]{}

\begin{document}

\author{Eric S. Tellez\thanks{CONACyT Consejo Nacional de Ciencia y Tecnolog\'ia, Direcci\'on de C\'atedras, Insurgentes Sur 1582, Cr\'edito Constructor 03940, Ciudad de M\'exico, M\'exico.}~\thanks{INFOTEC Centro de Investigaci\'on e Innovaci\'on en Tecnolog\'ias de la Informaci\'on y Comunicaci\'on, Circuito Tecnopolo Sur No 112, Fracc. Tecnopolo Pocitos II, Aguascalientes 20313, M\'exico.} \\ \url{eric.tellez@infotec.mx}
\and Daniela Moctezuma\footnotemark[1]~\thanks{Centro de Investigaci\'on en Geograf\'ia y Geom\'atica ``Ing. Jorge L. Tamayo'', A.C. Circuito Tecnopolo Norte No. 117, Col. Tecnopolo Pocitos II, C.P. 20313,. Aguascalientes, Ags, M\'exico.} \\ \url{dmoctezuma@centrogeo.edu.mx}
\and Sabino Miranda-Jim\'enez\footnotemark[1]~\footnotemark[2] \\ \url{sabino.miranda@infotec.mx}
\and Mario Graff \footnotemark[1]~~\footnotemark[2]\\ \url{mario.graff@infotec.mx}}

\title{An Automated Text Categorization Framework based on Hyperparameter Optimization}
\date{April. 2017}

\maketitle
\begin{abstract}

A great variety of text tasks such as topic or spam identification, user profiling, and sentiment analysis can be posed as a supervised learning problem and tackle using a text classifier. A text classifier consists of several subprocesses, some of them are general enough to be applied to any supervised learning problem, whereas others are specifically designed to tackle a particular task, using complex and computational expensive processes such as lemmatization, syntactic analysis, etc. Contrary to traditional approaches, we propose a minimalistic and wide system able to tackle text classification tasks independent of domain and language, namely $\mu$TC. It is composed by some easy to implement text transformations, text representations, and a supervised learning algorithm. These pieces produce a competitive classifier even in the domain of informally written text. We provide a detailed description of $\mu$TC along with an extensive experimental comparison with relevant state-of-the-art methods.  $\mu$TC was compared on 30 different datasets. Regarding accuracy, $\mu$TC obtained the best performance in 20 datasets while achieves competitive results in the remaining 10. The compared datasets include several problems like topic and polarity classification, spam detection, user profiling and authorship attribution. Furthermore, it is important to state that our approach allows the usage of the technology even without knowledge of machine learning and natural language processing.

\end{abstract}

\section{Introduction}
\label{sec:introduction}
Due to the large and continuously growing volume of textual data, automated text classification methods have taken an increasing interest of research community. Although many efforts have been proposed in this direction, it remains as an open problem. The arrival of massive data sources, like micro-blogging platforms, introduces new challenges where many of the prior techniques failed. Among the new challenges are: the volume and noisy nature of the data, the shortness of the texts that implies little context, the informal style also plagued of misspellings and lexical errors, among others.

These new data sources have made popular tasks such as {\em sentiment analysis} and {\em user profiling}. The sentiment analysis problem consists in determining the polarity of a given text, which can be a global polarity (about the whole text) or about a particular subject or entity. The user profiling task consists in, given a text, predicting some facts about the author, like her/his demographic information (e.g., gender, age, language or region). Such is the importance of these problems that in the research community several international competitions have been carried out in recent years. For example SemEval\footnote{http://alt.qcri.org/semeval2017/}, TASS\footnote{http://www.sepln.org/workshops/tass/2016/tass2016.php} and SENTIPOLC\footnote{http://www.di.unito.it/~tutreeb/sentipolc-evalita16/} are challenges for sentiment classifiers for Twitter data in English, Spanish, and Italian languages, respectively. PAN\footnote{http://pan.webis.de/} also opens calls for author profiling systems for English, Spanish and German languages. These problems are closely related to traditional text classification applications such as {\em topic classification} (e.g, classifying a news-like text into sports, politics, or economy), {\em authorship attribution} (e.g., identifying the author of a given text) and {\em spam detection}.

Usually, each of aforementioned problems is treated in a particular way, i.e., a method is proposed to solve adequately one classification task. Traditionally, this approach cannot  generalize to other related task, and, consequently, the methods are dependent on the problem; however, it is worth to mention that this specialization produces a lot of insight about the problem's domain. Conversely, in this contribution, we proposed a framework to create a text classifier regardless of both the domain and the language and based only a training set of labeled examples.

The idea of creating a text classifier almost independent of the language and domain is not novel, in fact, in our previous work \cite{B4MSA}, we introduced a combinatorial framework for sentiment analysis. There, aspects of language were considered  such as stopwords and tokenizers with special attention to lexical structures for negations. Also, particularities of the domain like {\em emoticons} and {\em emojis} are considered. The presented manuscript is a generalization and formalization of our previous work; this allows us to simplify the entire framework to work independently of both the language and the particular task, and empower the use of more sophisticated text treatments whenever it is possible and necessary.

As stated above, we tackle the problem of creating text classifiers that work regardless of both the domain and the language, with nothing more than a training set to be learned. The general idea is to orchestrate a number of simple text transformations, tokenizers, a set of weighting schemes, along with a Support Vector Machine (SVM) as classifier to produce effective text classification. More detailed, we look at the problem of creating effective text classifiers as a combinatorial optimization problem; where there is a search space containing all possible combinations of different text transformations, tokenizers, and weighting procedures with their respective parameters, and, on this search space, a meta-heuristic is used to search for a configuration that produces a highly effective text classifier. This model selection procedure is commonly named in the literature as {\em hyper-parameter optimization}. To emphasize the simplicity of the approach, we named it {\em micro} Text Classification or simply $\mu$TC.

This manuscript is organized as follows. The related work is presented in Section \ref{sec:relatedwork}. Section \ref{sec:combinatorialframework} describes our contribution in depth. In Section \ref{sec:setup}, all the experimental details are described. In Section \ref{sec:results}, we show an extensive experimental comparison of our approach with the relevant state-of-the-art methods over 30 different benchmarks. Finally, the conclusions are listed in Section \ref{sec:con}.


\section{Related work}
	\label{sec:relatedwork}
    

Let us start by describing a typical text classifier which can be summarized as a set of few, but complex, parts~\cite{khan2010review}. Firstly, the input text is passed to a lexical analyzer that both parses and normalizes the text, it outputs a list of tokens that represent the input text. The lexical analyzer typically includes some simple transformation functions like the removal of diacritic symbols and lower casing the text, but it also can make use of sophisticated techniques like stemming, lemmatization, misspelling correction, etc. Whereas, the tokens are commonly represented by words, pairs or triplets of adjacent words (bigrams or trigrams), and in general, sequences of words (word n-grams). It is also possible to extend this approach to sequences of characters (character n-grams). When it is allowed to drop the middle words of word n-grams, we obtain skip-grams. The usage of these techniques is driven by the human knowledge of the particular problem being tackled. Also, it is worth to mention that the entire process is tightly linked to the input language.

Secondly, the output of the lexical analyzer is commonly used to create high dimensional vectors where each token of the vocabulary has a corresponding coordinate in the vector. So, the value of each coordinate is associated with the weight of that token. The traditional way of weighting is to use the local and global statistics of tokens, popular examples of this approach are {\em TF}, {\em IDF}, {\em TFIDF}, and {\em Okapi BM25}; alternatively, some information measures like the entropy are commonly used as weight terms. Many times it is desirable to reduce the dimension of the vector space, and several techniques can be used for that purpose, just like {\em PCA}~\cite{PCA1987} (Principal Component Analysis), and {\em LSI}~\cite{LSI1998} (Latent Semantic Indexing).

Finally, the output of the weighting scheme, is used to create a training set which can be learned by a classifier. A classifier is a machine learning algorithm that learns the instances of a training set $\mathbb T$. In more detailed, the training set is a finite number of inputs and outputs, i.e., $\mathbb T = \{ (x_i, y_i) \mid i=1 \ldots n \}$ where $x_i$ represents the $i$-th input, and $y_i$ is the associated output. The objective is to find a function $\psi$ such that $\forall_{(x, y) \in \mathbb T}~\psi(x) = y$ and that could be evaluated in any element $x$ of the input space. In general, it is not possible to find a function $\psi$ that learns $\mathbb T$, perfectly. Consequently, a good classifier finds a function $\psi$ that minimizes an error function or maximizes a score function.

Perhaps, one of the first generic text classifier was proposed by Rocchio~\cite{rocchio1971relevance} that works by generating object prototypes based on centroids of a Voronoi partition over {\em TFIDF} vectors. This strategy shows the effort to reduce the necessary memory to fit in the hardware available at that time. Rocchio uses the nearest neighbor classifiers over prototypes to perform the predictions, the preprocessing of the text was left to the expertise of the user. Rocchio was the baseline and the study object in the area for a long time; such is the case of the work presented by Joachims~\cite{joachims1996probabilistic}, which describes a probabilistic analysis of the Rocchio algorithm.

With the purpose of improving the quality of the text classification task, Cardoso~\cite{cardoso2007improving} proposes the use of centroids to enhance the power of several typical classifiers, such as kNN (k-nearest neighbors) and SVM (Support Vector Machines). Also, Cardoso published a number of datasets in various preprocessing stages, which are popular among the text classification community because using them allows focusing on the weighting and classification algorithms, avoiding to tackle the text processing problem.

In \cite{androutsopoulos2004learningnuevo}, machine learning is used to create a spam detector. The proposed method uses a combination of a set of features, preprocessing steps or setup details, such as using lemmatization or not, using stop-list or not, keywords patterns, varying the length of the training corpus, etc. A similar work is presented by Androutsopoulos {\em et al}. in~\cite{Androutsopoulos-pus}. 

In the topic classification task, \cite{Debole2004} presents an experimental scheme with the Reuters dataset and three machine learning methods (Rocchio algorithm, k-NN, and SVM), and also, three-term selection functions (information gain, chi-square and gain ratio). \cite{hingmire2013document} proposes a topic modelling algorithm based on Latent Dirichlet Allocation (LDA) which assign one topic to an unlabeled document. Also, a combination of LDA and Expectation-Maximization (EM) algorithm is proposed.

Another approach to text classification is to move the focus from text processing and text classification, to improve the term-weighting; this is a successful strategy followed by recent works. Cummins~\cite{Cummins2006} proposes a method based on Genetic Programming to determine and evaluate several term weighting schemes for the vector space model. Escalante et al.~\cite{Escalante2015} present an approach to improve the performance of classical term-weighting schemes using genetic programming. Their approach outperforms standard schemes, based on an extensive experimental comparison. The authors also compare the Cummins~\cite{Cummins2006} approach over their benchmarks.

Lai et al.~\cite{lai2015recurrent} use both recurrent and convolutional neural networks to produce a term-weighting scheme that captures semantics from the text. Similarly to word embeddings~\cite{pennington2014glove,mikolov2013distributed}, the authors represent words based on their context and, also, they use skip-grams for text representation. The experimental results show higher values of macro-F1 in comparison to other state-of-the-art methods.

Vilares et al.~\cite{Vilares2017} introduce an unsupervised approach for multilingual sentiment analysis driven by syntax-based rules; the words are weighted based on the analysis of syntax-graphs. The authors provide experimental support for English, Spanish, and German. However, to support an additional language, it needs to implement several rules and a proper syntax parser.

Mozeti\v{c} et al.~\cite{multilingual2016} study the effect of the agreement among human taggers in the performance of sentiment classifiers. In this way, they compare several classifiers over a traditional text normalization and a vector representation with {\em TFIDF} weighting.They provide 14 tagged datasets for European languages; we selected some of them for our benchmarks. See Section \ref{sec:results} for more details.

Author profiling is another important task related to text categorization, where several advances have been proposed. In \cite{Lopez2015} the authors report their approach to perform author profiling; in particular, they describe the best classifier of the PAN'13 contest that consists on a distributional word representation based on the membership to each class along with a number of text standard text preprocessing, see \cite{pan2013}. Recently, in PAN'17~\cite{pan2017Overview}, some current works related to user profiling are presented. In this case, user profiling is related to gender and language-region classification. In this aspect, in \cite{basile2017}, an SVM, with linear kernel, in combination with word unigrams, character 3- to 5-grams and POS features are employed. In \cite{Martinc2017} the features were selected as word and POS n-grams, the number of emojis in the text, document sentiment, character flooding (counting the number of times that three or more identical character sequence appears in the text). Finally, a lexicon of important word is also employed.



\section{$\mu$TC: A Combinatorial Framework for Text Classification}
\label{sec:combinatorialframework}

Our approach consists in finding a competitive text classifier for a given task among a (possibly large) set of candidates classifiers. A text classifier is represented by the parameters that determine the classifier's functionality along with the input dataset. The search of the desired text classifier should be performed efficiently and accurately, in the sense that the final classifier should be competitive concerning the best possible classifier in the defined space of classifiers.

In the first part of this section, we will describe the structure of our approach, that is, we state the parameters defining our configuration space. Then, we define the $\mu$TC graph, which is the core structure used by the meta-heuristics implemented to find a good performing text classifier for a given task. In the road, we also describe the \textsf{score} function that encapsulates the functionality of the classifier and provides a numerical output necessary to maximize the efficiency of the classifier.


\subsection{The configuration space}
\label{ss/space}
As mentioned previously, a text classifier consists of well differentiated parts.
For our purposes, a classifier has the following parts: i) a list of functions that normalize and transform the input text to the input of tokenizers, ii) a set of tokenizer functions that transform the given text into a multiset of tokens, iii) a function that generates a vector from the multiset of tokens; and finally, iv) a classifier that knows how to assign a label to a given vector.
These pieces define a $\mu$TC space of configurations, which is defined by the tuple
	$(\mathcal{T}, \mathcal{G}, \mathcal{H}, {\Psi})$. In the following paragraphs a more detailed description is given.

	\begin{enumerate}
		\item $\mathcal{T} = \{T_i\}$ is the space of transformation functions, where $T_i$ is defined as the identity function $I$ and a set of related functions, mutually exclusive.\footnote{The identity function is defined as $I(S) = S$.}
We define the function $f(S) = (f_{|\mathcal{T}|}\circ \cdots\circ f_1)(S)$ such that $f_i \in T_i$, where the parameter $S$ is a text, i.e., a string of symbols.

		\item $\mathcal{G} = \{G_i\}$ is the set of tokenizer functions.
		Each $G_i$ is defined as either a function that returns $\emptyset$ or a simple tokenizer function, i.e., a tokenizer function is a function that extracts a list of subsequences of the given argument. More precisely, the function $g(S) = g_1(S) \cup \cdots \cup g_{|\mathcal{G}|}(S)$ is defined; where $g_i$ such that $g_i \in G_i$, extracts a list of subsequences of $S$. The final multiset is named as {\em bag of tokens}.
		\item $\mathcal{H}$ is a set of functions that transform a bag of tokens $v$ into a vector $\vec{v}$ of dimension $d$,  i.e., $h: \{S\}^+ \rightarrow \mathbb{R}^d$ where $S$ is a non empty string, $h \in \mathcal{H}$.
		The proper value of each vector's coordinate is also determined by $h$; the later task is commonly known as {\em weighting scheme}.

		\item Finally, $\Psi$ is a set of functions that create a classifier for a given labeled dataset as knowledge source.
	\end{enumerate}

	Now, let $\mathcal{C}$ be the set of all possible configurations of the $\mu$TC space; therefore, it is defined as follows:
	\[ \mathcal{C} =
	T_1 \times \cdots \times T_{|\mathcal{T}|} \times
	G_1 \times \cdots \times G_{|\mathcal{G}|} \times
	\mathcal{H} \times \Psi
	\]
	then, the size of $\mathcal{C}$ is described by
	\[|\mathcal{C}| =
	\left(\prod\limits^{|\mathcal{T}|}_{i=1}{|T_i|} \right) \cdot  2^{|\mathcal{G}|} \cdot
	|\mathcal{H}| \cdot |\Psi| \]

	Without loss of generality, the size of the search space can be summarized as
	$(2+O(1))^{|\mathcal{T}| + |\mathcal{G}|} \cdot |\mathcal{H}| \cdot |{\Psi}|$, where the $O(1)$ term captures the effect of $T_i$s with more than two member functions. This means that $|\mathcal{C}|$ is lower bounded by $2^{|\mathcal{T}| + |\mathcal{G}|}$, i.e., all $T_i$s are binary and both $\mathcal{H}$ and $\Psi$ are singletons.
	Even on the simplest setup, the configuration space grows exponentially with the number of possible transformations and tokenizers. Thus, in order to find the best item, it is necessary to evaluate the entire space; this is computationally not feasible.\footnote{For instance, evaluating each configuration takes about 10 minutes on a commodity workstation; more about this will be detailed in the experimental section.} A typical configuration space can contain billions of configurations such that the exhaustive evaluation is not feasible in current computers. To remain as a practical approach, instead of performing an exhaustive evaluation of $\mathcal{C}$ to find the best configuration, we soften the problem to find a (very) competitive configuration;
    then it can be solved as a combinatorial optimization problem, in particular, as the maximization of a \textsf{score} function.

\subsection{The configuration graph}
\label{ss/graph}
	In order to solve the combinatorial problem with local search-based meta-heuristics, it is necessary to create a graph where the vertex set corresponds to $\mathcal{C}$, and the edge set corresponds to the neighborhood of each vertex, $\{N(c) \subseteq \mathcal{C}^+ \mid c \in \mathcal{C}\}$. The edges are simply denoted by the neighborhood function $N$, so $(\mathcal{C}, N)$ is a $\mu$TC graph.

	Our main assumption is simple and feasible, the function \textsf{score} slowly varies on similar configurations, such that we can assume some degree of {\em locally concaveness}, in the sense that a well-performing local maximum can be reached using greedy decisions at some given point. Even when this is not true in general, the solver algorithm should be robust enough to get a good approximation even when the assumption is valid only with some degree of certainty.
	To induce the search properties, the neighborhood $N$ should be defined in such a way that neighborhoods describe only similar configurations.
	For this matter, we should define a distance function between configurations. First, we must define a comparison function,
	\begin{equation}
	\Delta(a, b) = \begin{cases}
	1 & \text{if } a \text{ and } b \text{ are the same function} \\
	0 & \text{otherwise}
	\end{cases}
	\end{equation}
	Since each configuration is a tuple of functions, the Hamming distance over configurations is naturally defined as follows
	\begin{equation}
	d_H(u, v) = \sum^{|\mathcal{T}|+|\mathcal{G}|+2}_{i=1} \Delta(u_i, v_i).
	\end{equation}
	Now, we can define $N(c, r_\textsf{max}) = \{u \in \mathcal{C} \mid 0 < d_H(u, c) \leq r_\textsf{max}\}$, for any $r_\textsf{max}$ and a configuration $c$. However, the number of items grows exponentially with the radius, and therefore, the notion of locality will be rapidly degraded. To maintain the locality, we define  the neighborhood as:
	\begin{equation}
	N(c) = \{u \in \mathcal{C} \mid d_H(c, u) = 1\}.
	\end{equation}

	Under this construction scheme, the diameter of $(\mathcal{C},N)$ is determined by the length of the configuration tuple, i.e., $O(\log{|\mathcal{C}|})$, the diameter determines the number of hops in the $\mu$TC graph that an optimal \textsf{opt} algorithm will perform, in the worst case. However, since the best configuration is unknown, we must use {\em score} as an {\em oracle} that leads our navigation at each step.

\subsection{The \text{score} function}
    The \textsf{score} function evaluates the performance of the text classifier defined by the configuration with the given training and test sets. Without loss of generality, the evaluation of a configuration $c \in \mathcal{C}$ can be described by three main steps:

\begin{enumerate}
    \item The dataset $\mathcal{D}$ is divided into $D_\text{train}$ and $D_\text{test}$.
    \item The $\mu$TC algorithm described by $c$ learns from $D_\text{train}$.
    \item The prediction performance of $c$ is evaluated using the dataset $D_\text{test}$, more details are given below.
\end{enumerate}
These steps can be modified to support cross-validation, schemes like $k$-folds or bagging, which provide a more robust way to measure the performance of a classifier. The details of these measurement strategies are beyond the scope of this manuscript, the interested reader is referenced to Ch. 9 of \cite{Olson2008}.

Now, please recall from \S\ref{ss/space} that $c$ contains the parameters for a number of functions that transform the input text into its associated label. Given a configuration $c$, a classifier $\psi$ is created using the labeled dataset transforming all texts in the training set to its corresponding vector form, i.e., $h \circ g \circ f(S)$ for $S\in D_\text{train}$.
Once the classifier $\psi$ is trained, the associated label for all $S \in D_\text{test}$ is computed as $\psi \circ h \circ g \circ f (S)$. Finally, the performance of $c$ is computed comparing the predicted labels against the actual ones; a typical \textsf{score} function will use F1 (macro or micro), accuracy, precision, or recall, to measure the quality of the text classifier.

\subsection{Optimization process}
\label{ss/search}
The core idea to solve the optimization problem is to navigate the graph $(\mathcal{C}, N)$ using a combination of two meta-heuristics. In the following paragraphs, we briefly review the techniques we used to solve the combinatorial problem, a proper survey of the area is beyond the scope of this manuscript. However, the interested reader is referred to \cite{burke2005search, battiti2008reactive}.

To maintain $\mu$TC in practical computational requirements, we select two types of fast meta-heuristics, {\em Random Search} \cite{bergstra2012random} and {\em Hill Climbing} \cite{burke2005search,battiti2008reactive} algorithms.
    The former consists in selecting the best performing configuration among the set $\mathcal{C}'$ randomly chosen from $\mathcal{C}$, that is,
    $$\arg\max_{c \in \mathcal{C}'} \textsf{score}(c),$$
where the size of $\mathcal{C'}$ is an open parameter dependent on the task.
On the other hand, the core idea behind Hill Climbing is to explore the configuration's neighborhood $N(c)$ of an initial setup $c$ and then greedily update $c$ to be the best performing configuration in $N(c)$. The process is repeated until no improvement is possible, that is, $$\textsf{score}(c) \geq \max_{u \in N(c)} \textsf{score}(u).$$ We improve the whole optimization process applying a Hill Climbing procedure over the best configuration found by a Random Search.
    We also add memory to avoid a configuration to be evaluated twice.\footnote{In principle, this is similar to Tabu search; however, our implementation is simpler than a typical implementation of Tabu search.}

	Summarizing, the optimization process is driven by the tuple $(\mathcal{C}, \mathcal{D}, \mathsf{score}, \mathsf{opt})$, where i) $C$ is the $\mu$TC space, ii) $\mathcal{D}$ means the training set of labeled texts, iii) $\mathsf{score}$ is the function to be maximized, and finally, iv) $\mathsf{opt}$ is a combinatorial optimization algorithm that uses $\mathsf{score}$ and $\mathcal{D}$ to find an almost optimal configuration in $\mathcal{C}$.

\section{Experimental setup}
\label{sec:setup}
This section describes the general setup used to characterize and compare our method with the related state-of-the-art. In particular, we define the set of functions used to create our $\mu$TC space; and also, we detail the benchmarks used in the comparison.

All the experiments were run in an Intel(R) Xeon(R) CPU E5-2640 v3 @ 2.60GHz with 32 threads and 192 GiB of RAM running CentOS 7.1 Linux.
We implemented $\mu$TC\footnote{Available under Apache 2 license at \url{https://github.com/INGEOTEC/microTC}} on Python.
    To characterize the performance of $\mu$TC and compare it to the relevant state of the art, we selected a number of popular benchmarks in the literature; these datasets are described below. It is worth to mention that we bias our selection to benchmarks coming from popular international challenges.
    With the purpose of avoiding over-fitting, we performed the model selection using \textsf{score} as a $3$-fold cross-validation of the specified performance measure, see Table~\ref{tab:database-description}.
    We decided to use cross-validation for this stage because we observed over-fitting for small datasets, like those found in authorship attribution, when we use a static train-test partitions to perform model selection.
    A brief experimental study of the effect of the validation schemes is presented in \S\ref{sec:cross-validation-study}.

\subsection{About our particular $\mu$TC space}
As state before, $\mu$TC is a framework to create text classifiers searching for best models in a configuration space. This space can be adjusted for any particular problem, but here, we consider a general enough space to match a disparity of benchmarks (listed below in \S\ref{sec:benchmarks}).

When the knowledge about the domain is low, then a large and generic configuration space should be used. It could be tempting to learn about the domain using the information found by the optimization process; this is clearly possible. However, it is encouraged to take into account that the search process will take decisions to match the particular dataset, not the domain, and any generalization of the knowledge must be curated by an expert in the domain. It is important to mention that large configuration spaces will consume a lot of computational time to be optimized.

On the other hand, a hand-crafted configuration space for a given problem can yield to very fast processing times; however, a vast knowledge of the domain is required to reach this state. In this case, we discard the possibility of discovering new knowledge on the domain and take advantage of the particularities of the dataset that a more general configuration space can provide.

To tackle with the disparate list of benchmarks, we select a generic large configuration space defined in the following paragraphs.

\paragraph{Preprocessing functions $\mathcal{T} = \{T_1, \dots, T_{|\mathcal{T}|}\}$} We associate $T_i$ to the following function sets.
    \begin{description}
    \item[hashtag-handlers.] Defined as $\{remove\_htags, group\_htags, identity\}$, the idea is to allow to remove or group into a single tag all hash tags, for $remove\_htags$ and $group\_htags$, respectively; the $identity$ function lets the text unmodified. The format of a hash tag is that introduced by Twitter $\#words$, but now popular along many data sources.
    \item[number-handlers.] Defined as $\{remove\_num, group\_num, identity\}$, this function set contains functions to remove, group, or left untouched numbers in the text.
    \item[url-handlers.] Defined as $\{remove\_urls, group\_urls, identity\}$, this function set contains functions to remove, group, or left untouched numbers in the text.
    \item[usr-handlers.] Defined as $\{remove\_usr, group\_usr, identity\}$, this function set contains functions to remove, group, or left untouched users and host domains in the text. The pattern being tackled is \texttt{@user} this is a popular way to denote users in several social networks; the pattern also matches naturally with the domain part of email addresses.
    \item[diacritic-removal.] Defined as $\{remove\_diac, identity\}$, this function set contains functions to remove, or left untouched, diacritic symbols in the text. The objective is to reduce composed symbols like \texttt{\'a,\"a,\~a,\^a,} or \texttt{à} to simply \texttt{a}.  This is a well known source of errors in informal text written in languages making hard use of  diacritic symbols
    \item[duplication-removal.] Defined as $\{remove\_dup, identity\}$, this function set contains functions to remove, or left untouched, duplicated contiguous symbols in the text.
    \item[punctuation-removal.] Defined as $\{remove\_punc, identity\}$, this function set contains functions to remove, or left untouched, duplicated punctuation symbols in the text. The list of punctuation symbols includes several symbols like \texttt{;,:,.,-,',",(,),[,],\{,\},$\sim$,<,>,?,!,} among others.
    \item[lower-casing.] Defined as $\{lower\_case, identity\}$ contains functions to normalize the case of the text or left untouched.
    \end{description}

\paragraph{The list of tokenizers $\mathcal{G} = \{G_1, \dots, G_{|\mathcal{G}|}\}$} After all text normalization and transformation, a list of tokens should be extracted. We use three schemes for our tokenizers.
    \begin{description}
    \item[Word n-grams.] This family of tokenizers firstly tokenizes the text into words, and then, produces $m-n+1$ tokens for $m$ words, i.e. word $n$-grams. An $n$-gram  is a string of $n$ consecutive words. For example, ``\textsf{The red car is in front of the tree}'' creates the following 3-grams: \textsf{The red car, red car is, car is in, is in front, in front of, front of the, of the tree}.
    \item[Character n-grams.] This family of tokenizers does not assume anything about the text and splits the input text to all $n$-sized substrings, i.e., $m-n+1$ substrings of characters for a text of $m$ characters. For example, the character 4-grams of ``\textsf{I like the red car}'' are \textsf{I\_li, \_lik, like, ike\_, ke\_t, e\_th, \_the, the\_, he\_r, e\_re, \_red, red\_, ed\_c, d\_ca, \_car}. We use the symbol \textsf{\_} to show the symbol space.
    \item[Skip-grams.] Skip-grams are similar to word n-grams but allowing to {\em skip} the middle parts. For example, the $(2,1)$ skip-grams\footnote{Two words, skipping one in the middle} of the previous example are \textsf{I-the, like-red, the-car}. The idea behind this family of tokenizers is to capture the occurrence of related words that are separated by some unrelated words.
    \end{description}
    For this matter, instead of selecting one or another tokenizer scheme, we allow to select any of the available tokenizers, and perform the union of the final multisets of tokens. For instance, our configuration space considers three word n-grams tokenizers ($n=1,2,3$), nine character n-grams ($n=1..9$), and three skip-grams $(3,1), (2,2), $ and $(2,1)$.

\paragraph{Weighting schemes $\mathcal{H}$} After we obtained a multiset (bag of tokens) from the tokenizers, we must create a vector space. We selected a small set of frequency filters and the TFIDF scheme to weight the coordinates of the vector.
On one hand, we consider a sequential list of filters \textsf{max-filter} and \textsf{min-filter}, and then, we select to use the term frequency (TF) or the TFIDF as weight. For the \textsf{max-filter} we delete all tokens surpassing the frequency threshold of $\alpha\textsf{max-freq}$, where \textsf{max-freq} is the maximum frequency in of a token in the collection. We consider four filters, for instance we use $\alpha \in \{0.9, 0.95, 0.99, 1.0\}$.
For the \textsf{min-filter} we delete all tokens not reaching the frequency threshold of $freq$, for instance we use, $freq \in \{1, 3, 5, 10\}$. Notice that  $\alpha = 1.0$ and $freq = 1$ does not perform any filtering. So, we have embedded 32 different configurations for weighting.

\paragraph{Classifier $\Psi$} We decide to use a singleton set populated with an SVM with a linear kernel. It is well known that SVM performs excellently for very large dimensional input (which is our case), and the linear kernel also performs well under this conditions. We do not optimize the parameters of the classifier since we are pretty interested in the rest of the process. We use the SVM classifier from {\em liblinear}, Fan et al.~\cite{fan2008liblinear}.

\paragraph{On the final configuration space} The task of finding the best model for the space of configurations is hard.
The number of possible configurations of $\mathcal{F}$ is $1296$ (i.e., four trivalent functions sets and four bivalent function sets). From the above configuration, the number of possible tokenizers is 81; also, we have 32 different weighting combinations. So, the configuration is space contains more than 3.3 million configurations. For instance, a configuration needs close to ten minutes to be evaluated, i.e., a sentiment analysis benchmark with ten thousand tweets. Therefore, an exhaustive evaluation of the configuration space will need up to 64 years. Even implementing it in a large distributed cluster the process needs too much time to complete. Such power of computing is not easily accessible.
Nonetheless, if we soften the problem to finding not the best model but an excellent one, we can use an algorithm for combinatorial optimization, as explained in \S\ref{sec:combinatorialframework}.

	\subsection{On the preparation of the input text}
	Since $\mu$TC considers the preprocessing step among its parts, we tried to collect all datasets in raw text, without any kind of preprocessing transformations.
	This was not possible in the general case, mostly due to the aging of datasets; we consider the following text preparation states, in the style of Cachopo~\cite{cardoso2007improving}:
	\begin{itemize}
		\item the {\em raw} text corresponds to the original, non-formatted text
		\item the {\em all-terms} converts all text into lowercase, also, all diacritic symbols and punctuation marks are removed, and all spacing symbols are normalized to a single space
		\item the {\em no-short} dataset removes all terms having less or equal than three characters
		\item the {\em no-stopwords} dataset also removes all non discriminant words for English (adjetives, adverbs, conjunctions, articles, etc.)
		\item finally, after the previous steps, all words were transformed by the Porter's stemmer for English~\cite{NLTK2009} to generate the {\em stemmed} dataset.

	\end{itemize}

	For instance, we use the {\em all-terms} for \textsf{R8}, \textsf{R10}, \textsf{R52} and \textsf{WebKB}; for \textsf{CADE} we use the {\em stemmed} version. In these cases, we used the datasets prepared by Cachopo~\cite{cardoso2007improving}.
	In other cases, we use the raw text. The effect of using one or another state is studied in Section \ref{sec:input-stage}.

\subsection{Benchmark description}
\label{sec:benchmarks}
     The text classification literature has a myriad of datasets, performance measures, and validation schemes.
    We select several prominent and popular benchmark configurations in the literature; for instance, we select to work with topic classification, spam identification, author profiling, authorship attribution, and sentiment analysis.
    To avoid implementation mistakes, we directly use the reported performances by the literature; nevertheless, we are restricted to compare under the same circumstances.
    Table~\ref{tab:database-description} describes the language and number of classes of each dataset; it also describes the kind of validation; in particular, we consider two validation schemes:
    i) 10-fold cross-validation, and ii) a static train-test partition of the specified sizes. The diversity of benchmarks and validation schemes help us to prove the functionality of our approach in many circumstances.

	\begin{table}
		\caption{Description of the benchmarks and its associated performance measure}
		\label{tab:database-description}
        \centering
		\resizebox{!}{.5\textheight}{
			\begin{minipage}{\textwidth}
				\begin{tabular}{clrr rrc}
					\toprule
					name  & language & \multicolumn{3}{c}{\#documents } & \#classes & performance \\ \cline{3-5}
					&          & total              & train & test &            & measure \\
					\midrule \multicolumn{7}{c}{Topic Classification} \\ \midrule
					R8      & English    & 7,674   & 70\%  & 30\% & 8  & macro-F1 \\
					R10     & English    & 8,008   & 70\%  & 30\% & 10 & macro-F1 \\
					R52     & English    & 9,100   & 70\%  & 30\% & 52 & macro-F1 \\
					News-4  & English    & 13,919  & 70\%  & 30\% & 4  & macro-F1 \\
					News-20 & English    & 20,000  & 70\%  & 30\% & 20 & macro-F1 \\
					WebKB   & English    & 4,199   & 70\%  & 30\% & 4  & macro-F1 \\
					CADE    & Portuguese & 40,983 & 70\%  & 30\% & 12 & macro-F1 \\
					\midrule \multicolumn{7}{c}{Spam Identification} \\ \midrule
					Ling-Spam & English        & 2,893  & \multicolumn{2}{r}{--- 10-fold ---} & 2 & macro-F1 \\
					PUA       & English$^\dag$ & 1,142  & \multicolumn{2}{r}{--- 10-fold ---} & 2 & macro-F1 \\
					PU1       & English$^\dag$ & 1,099  & \multicolumn{2}{r}{--- 10-fold ---} & 2 & macro-F1 \\
					PU2       & English$^\dag$ & 721    & \multicolumn{2}{r}{--- 10-fold ---} & 2 & macro-F1 \\
					PU3       & mixed$^\dag$   & 4,139  & \multicolumn{2}{r}{--- 10-fold ---} & 2 & macro-F1 \\
					\midrule
                    \multicolumn{7}{c}{Author Profiling} \\ \midrule
                    \multirow{2}{*}{\shortstack[c]{PAN'13 Gender  \& \\ Age group}}
					  & English & 242,040 & 236,600 & 25,440 & 2 \& 3      & accuracy \\
					~ & Spanish & 84,060  & 75,900  & 8,160  & 2 \& 3      & accuracy \\
                    \hline
                    \multirow{4}{*}{\shortstack[c]{PAN'17$^\ddag$ Gender \& \\ Language Variety}}
                      & Arabic     &   -   & 2,400 &   -   & 2 \& 4 & accuracy \\
                    ~ & English    &   -   & 3,600 &   -   & 2 \& 6 & accuracy \\
                    ~ & Spanish    &   -   & 4,200 &   -   & 2 \& 7 & accuracy \\
                    ~ & Portuguese &   -   & 1,200 &   -   & 2 \& 2 & accuracy \\
					\midrule \multicolumn{7}{c}{Authorship Attribution} \\ \midrule
					CCA      & English & 1,000   & 500   & 500  & 10     & macro-F1 \\
					NFL      & English & 97      & 52    & 42   & 3      & macro-F1\\
					Business & English & 175     & 85    & 90   & 6      & macro-F1\\
					Poetry   & English & 200     & 145   & 55   & 6      & macro-F1\\
					Travel   & English & 172     & 112   & 60   & 4      & macro-F1\\
					Cricket  & English & 158     & 98    & 60   & 4      & macro-F1\\
					\midrule \multicolumn{7}{c}{Multilingual Sentiment Analysis} \\ \midrule
					Arabic  & Arabic        &  2,000 & \multicolumn{2}{r}{--- 10-folds ---} & 3   & macro-F1 \\
					German  & German        & 91,502 & \multicolumn{2}{r}{--- 10-folds ---} & 3   & macro-F1 \\
					Portuguese & Portuguese & 86,062 & \multicolumn{2}{r}{--- 10-folds ---} & 3   & macro-F1 \\
					Russian & Russian       & 69,100 & \multicolumn{2}{r}{--- 10-folds ---} & 3   & macro-F1 \\
					Spanish & Spanish       & 19,767 & \multicolumn{2}{r}{--- 10-folds ---} & 3   & macro-F1 \\
					Swedish & Swedish       & 49,255 & \multicolumn{2}{r}{--- 10-folds ---} & 3   & macro-F1 \\
					\bottomrule
				\end{tabular}\\~\\
				$\dag$ these datasets are encoded in a way that the original text is loss, however it preserves the document's distribution.\\
				$\ddag$ here, the documents are Twitter's profiles, each user is described 100-300 single entries for a total of 1,265,898 tweets for all languages in the training set.
			\end{minipage}
		}
	\end{table}


	The Reuters-21578\footnote{\url{http://www.daviddlewis.com/resources/testcollections/reuters21578/}} is one of the most used collection for text categorization research. The documents were manually labeled by personnel from Reuters Ltd.
	%
	The 20Newsgroup\footnote{\url{http://people.csail.mit.edu/jrennie/20Newsgroups}} dataset is very popular in text classification area and it contains news related to different topics originally collected by Ken Lang.
	The WebKB dataset\footnote{\url{http://www.cs.cmu.edu/~webkb/}} contains university webpages. This dataset is composed of the webpages classified in seven different categories: student, faculty, staff, department, course, project and other. We use the four most popular classes in our experiments.
	The CADE dataset~\cite{cardoso2007improving} is another collection of webpages, specifically Brazilian webpages classified by human experts. This collection contains a total of 12 classes, e.g. services, sports, science, education, news, among others.
	The PU~\cite{Androutsopoulos-pus} is a collection of emails written in English and other languages, classified as spam and non-spam messages; this collection contains the following datasets: PUA, PU1, PU2 and PU3.
	%
	%
	Ling-Spam dataset~\cite{androutsopoulos2000evaluation} is also a spam dataset.
	%
	PAN contest~\cite{pan2013,pan2017Overview} has several tasks, between them are author identification and author profiling.
	The author profiling task is a forensic linguistics problem that consisnts in detecting gender and age for the author (PAN'13). For the PAN'17 age identification task was replaced by the task of determining the language variety of the writter, also, the number of different languages was increased to four. As listed in Table~\ref{tab:database-description}, the official dataset is undisclosed, and each algorithm must be evaluated with the {TIRA} evaluation platform.\footnote{\url{https://tira.io}} 
	The Authorship Attribution datasets \cite{Escalante2015} are a set of different types of topics: CCA, NFL, Business, Poetry, Travel and Cricket. The objective of these datasets is to determine the authorship of each document.
	The Multilingual Sentiment Analysis are a set of tweets in different languages: Arabic, German, Portuguese, Russian, Swedish and Spanish. The purpose of these datasets is classifying each tweet as negative, neutral, or positive polarity.

	  A detailed description of all these datasets is provided in Table \ref{tab:database-description}, where there can be found some particularities of the dataset like the written language, the number of documents, the kind of evaluation (independent train-test sets or $k$-folds), the number of classes, and the performance measure optimized by $\mu$TC.

\section{Experimental Results}
\label{sec:results}
This section is dedicated to comparing our work with the relevant state-of-the-art methods described above. Also, we characterize the generalization power in terms of the validation scheme.

The first task analyzed is authorship attribution, Table~\ref{tab:authorship} shows the macro-F1 and accuracy performances for a set of authorship attribution benchmarks. Here, we compare $\mu$TC with two term-weighting schemes \cite{Escalante2015} and \cite{Cummins2006}.
    The pre-processing stage of the $\mu$TC's input is {\em all-terms}; others use the {\em stemmed} stage. The best performing classifiers are created by $\mu$TC, except for \textsf{NFL} where alternatives perform better. In the case of \textsf{Business}, Escalante et. al~\cite{Escalante2015} performs slightly better only in terms of accuracy. Please notice that \textsf{NFL} and \textsf{Bussiness} are among the smaller dataset we tested, the low performance of $\mu$TC can be produced by the low number of exemplars, while alternative schemes take advantage of the few samples to compute better weights.

	\begin{table}[ht]
		\centering
		\caption{Authorship Attribution Data sets.}
		\label{tab:authorship}
		\begin{minipage}{0.5\textwidth}
			\resizebox{\textwidth}{!}{
				\begin{tabular}{lccc}
					& \multicolumn{3}{c}{\textbf{macro-F1}} \\
					Dataset & {Cummins~\cite{Cummins2006,Escalante2015}} & {Escalante \cite{Escalante2015}} & {$\mu$TC} \\ \cline{2-4}
					CCA     & 0.0182 & 0.7032 & {\bf 0.7633}  \\
					NFL     & {\bf 0.7654} & 0.7637 & 0.7422 \\
					Business& 0.7548 & 0.7808 & {\bf 0.8199} \\
					Poetry  & 0.4489 & 0.7003 & {\bf 0.7135} \\
					Travel  & 0.6758 & 0.7392 & {\bf 0.8621} \\
					Cricket & 0.9170 & 0.8810 & {\bf 0.9665} \\
				\end{tabular}
			}
		\end{minipage}\begin{minipage}{0.5\textwidth}
		\resizebox{\textwidth}{!}{
			\begin{tabular}{lccc}
				& \multicolumn{3}{c}{\textbf{Accuracy}} \\
				Dataset & {Cummins~\cite{Cummins2006,Escalante2015}} & {Escalante \cite{Escalante2015}} & {$\mu$TC} \\ \cline{2-4}
				CCA     & 0.1000 & 0.7372 & {\bf 0.7660}  \\
				NFL     & 0.7778 & {\bf 0.8376} & 0.7555 \\
				Business& 0.7556 & {\bf 0.8358} & 0.8222 \\
				Poetry  & 0.5636 & 0.7405 & {\bf 0.7272} \\
				Travel  & 0.6833 & 0.7845 & {\bf 0.8667} \\
				Cricket & 0.9167 & 0.9206 & {\bf 0.9667} \\
			\end{tabular}
		}
	\end{minipage}

\end{table}

In Table \ref{tab:authorship} the results of PAN'13 competition are presented. According to the contest report \cite{pan2013}, the best results were achieved by Pastor, Santosh, and Meina. In this benchmark, $\mu$TC produces the best result in all average cases. In a fine-grained comparison, only Meina surpasses $\mu$TC on the gender identification for English.

\begin{table}[!th]
	\centering
	\caption{Performance of $\mu$TC on the author profiling task of the PAN'13 competition; all values are the accuracy score in the specified subtask.}
	\label{tab:pan2013}
	\resizebox{0.68\textwidth}{!}{
    \begin{tabular}{lcccc}
        \toprule
                & Task   & English    & Spanish & Avg.\\
        \midrule
                & Age    & 0.6605     & 0.6897  & \bf 0.6751 \\
        $\mu$TC & Gender & 0.5867     & 0.6750  & \bf 0.6309 \\
                & Joint  & 0.3946     & 0.4587  & \bf 0.4267 \\
        \midrule
                & Age    & 0.6572     & 0.6558  & 0.6565 \\
    Pastor L.   & Gender & 0.5690     & 0.6299  & 0.5995 \\
                & Joint  & 0.3813     & 0.4158  & 0.3985 \\
        \midrule

                & Age    & 0.6408     & 0.6430  & 0.6419 \\
        Santosh & Gender & 0.5652     & 0.6473  & 0.6063 \\
                & Joint  & 0.3508     & 0.4208  & 0.3858 \\
        \midrule
                & Age    & 0.6491     & 0.4930  & 0.5711 \\
        Meina   & Gender & 0.5921     & 0.5287  & 0.5604 \\
                & Joint  & 0.3894     & 0.2549  & 0.3222 \\
        \bottomrule

    \end{tabular}
	}
\end{table}

Table \ref{tab:pan2017} shows the performance of $\mu$TC in the PAN'17 benchmark. The table also lists the best three results of the challenge, reported as statistically equivalent in \cite{pan2017Overview}, these works are detailed in \S\ref{sec:relatedwork}.
Please note that the result by Tellez et al.~\cite{tellez2017gender} was generated with $\mu$TC but using a special term-weighting scheme based on entropy instead of TFIDF (or TF). The details of the entropy based term-weighting scheme are beyond the scope of this contribution; the interested reader is referenced to \cite{tellez2017gender}.
The plain $\mu$TC, as described in this manuscript, achieves accuracies of 0.7880 and 0.8849, respectively for gender and variety identification. The joint prediction of both classes achieves an accuracy of 0.7038. These score values locate the plain $\mu$TC in the eighth position in the official rank, see \cite{pan2017Overview}.

\begin{table}[!th]
\centering
\caption{Author profiling: PAN2017 benchmark \cite{pan2017Overview}, all methods were scored with the official gold-standard. All scores are based on the accuracy computation over the specified subset of items.}
\label{tab:pan2017}
\resizebox{0.9\textwidth}{!}{
\begin{tabular}{cccc ccc}

\toprule

Method         & Task    & Arabic & English&Spanish & Portuguese     & Avg.    \\ \midrule
               & Gender  & 0.7569 & 0.7938 & 0.7975 & 0.8038 & 0.7880 \\  
$\mu$TC        & Variety & 0.7894 & 0.8388 & 0.9364 & 0.9750 & 0.8849 \\  
               & Joint   & 0.6081 & 0.6704 & 0.7518 & 0.7850 & 0.7038 \\  
\midrule
               & Gender  & 0.8006 & 0.8233 & 0.8321 & 0.8450 & \bf 0.8253 \\
Basile et al. \cite{basile2017}
               & Variety & 0.8313 & 0.8988 & 0.9621 & 0.9813 & \bf 0.9184 \\
               & Joint   & 0.6831 & 0.7429 & 0.8036 & 0.8288 & \bf 0.7646 \\
\midrule
               & Gender  & 0.8031 & 0.8071 & 0.8193 & 0.8600 & 0.8224 \\
Martinc et al. \cite{Martinc2017}
               & Variety & 0.8288 & 0.8688 & 0.9525 & 0.9838 & 0.9085 \\
               & Joint   & 0.6825 & 0.7042 & 0.7850 & 0.8463 & 0.7545 \\
\midrule
               & Gender  & 0.7838 & 0.8054 & 0.7957 & 0.8538 & 0.8097 \\
Tellez et al. \cite{tellez2017gender}
               & Variety & 0.8275 & 0.9004 & 0.9554 & 0.9850 & 0.9171 \\
               & Joint   & 0.6713 & 0.7267 & 0.7621 & 0.8425 & 0.7507 \\

\bottomrule
\end{tabular}
}
\end{table}

Table \ref{tab:topic} reports the performance over topic classification benchmarks. This experiments considered several {\em news} datasets.\footnote{Please refer to Table \ref{tab:database-description} for the detailed description of each benchmark.}
Our approach, $\mu$TC, reaches best results in most of the datasets with exception of News-20 and News-4 where $\mu$TC reaches second and third best performance.

\begin{table}[ht]
	\centering
	\caption{Topic Classification Datasets}
	\label{tab:topic}
	\begin{minipage}{\textwidth}
		\resizebox{\textwidth}{!}{
			\begin{tabular}{rccc cccc}
				& \multicolumn{7}{c}{\textbf{macro-F1}} \\
				& Reuters-8C & Reuters-10C & Reuters-52C & News-4C & News-20C & WebKB & CADE \\ \cline{2-8}
				Debole  \cite{Debole2004}
				&     -   &    -   &    -   &     -   &     -   &    -   &    -   \\
				Escalante \cite{Escalante2015}
				& 0.9135  & 0.9184 &    -   &     -   & 0.6797  & 0.8879 & 0.4103 \\
				Cummins \cite{Cummins2006,Escalante2015}
				& 0.8830  & 0.8759 &    -   &     -   & 0.6645  & 0.7197 &    -   \\
				Lai CNN \cite{lai2015recurrent}
				&     -   &    -   &    -   &  0.9479 &     -   &    -   &    -   \\
				Lai RNN \cite{lai2015recurrent}
				&     -   &    -   &    -   & {\bf 0.9649}&     -   &    -   &    -   \\
				Hingmire\cite{hingmire2013document}
				&     -   &    -   &    -   &     -   &     -   & 0.7190 &    -   \\
				Cachopo \cite{cardoso2007improving}
				&     -   &    -   &    -   &     -   &     -   &    -   &    -   \\
				$\mu$TC   & {\bf 0.9698}&{\bf 0.9662}&{\bf 0.6746}&  0.9432 &{\bf 0.8269}&{\bf 0.9098}&{\bf 0.5687}\\
			\end{tabular}
		}
	\end{minipage}

	\begin{minipage}{\textwidth}
		\resizebox{\textwidth}{!}{
			\begin{tabular}{rccc cccc}
				& \multicolumn{7}{c}{\textbf{accuracy}} \\
				& Reuters-8C & Reuters-10C & Reuters-52C & News-4C & News-20C & WebKB & CADE \\ \cline{2-8}
				Debole \cite{Debole2004}
				&    -   & 0.7040 &    -   &    -   &     -   &    -   &    -   \\
				Escalante \cite{Escalante2015}
				& 0.9056 & 0.8821 &    -   &    -   &  0.6623 & 0.8912 & 0.5380 \\
				Cummins \cite{Cummins2006,Escalante2015}
				& 0.7440 & 0.7659 &    -   &    -   &  0.6578 & 0.7542 &    -   \\
				Lai CNN \cite{lai2015recurrent}
				&    -   &    -   &    -   &    -   &     -   &    -   &    -   \\
				Lai RNN \cite{lai2015recurrent}
				&    -   &    -   &    -   &    -   &     -   &    -   &    -   \\
				Hingmire \cite{hingmire2013document}
				&    -   &    -   &    -   & 0.9360 &     -   &    -   &    -   \\
				Cachopo \cite{cardoso2007improving}
				& 0.9049 &    -   & 0.8482 &    -   &{\bf 0.8460}& 0.8300 & 0.5071 \\
				$\mu$TC   &{\bf 0.9214}&{\bf 0.9236}&{\bf 0.9376}&{\bf 0.9390}& 0.8348  &{\bf 0.9191}&{\bf 0.6174}\\
			\end{tabular}
		}
	\end{minipage}
\end{table}

In sentiment analysis task we compared the datasets reported in ~\cite{arabic2015,arabic2016}. Moreover, we reported the results obtained with the B4MSA approach~\cite{tellez2016simple}. B4MSA is a method for multilingual polarity classification considered as a baseline to build more complex approaches\footnote{https://github.com/INGEOTEC/b4msa}. It is important to note that from each dataset reported in ~\cite{arabic2015,arabic2016}, both approaches, B4MSA and $\mu$TC, use a subset specified in Table \ref{tab:sentimentanalysis}; e.g. in Arabic language we used 100\%, in German we used 80\% of the dataset and so on (all specified in table).

In Table \ref{tab:sentimentanalysis}, it can be seen that best results were obtained with B4MSA and $\mu$TC in all the cases, and both results are very close.

\begin{table}[!h]
	\caption{Multilingual sentiment analysis}
	\label{tab:sentimentanalysis}
	\centering
	\resizebox{0.7\textwidth}{!}{
		\begin{tabular}{r l c c  c}
			\toprule
			language       &  & macro-$F_1$ & accuracy \\
			\midrule \multirow{4}{*}{Arabic}
			& Salameh et al.~\cite{arabic2015} & -    & 0.787 \\
			& Saif et al.~\cite{arabic2016}    & -    &  0.794 \\
			& B4MSA (100\%)                    & 0.642 & \bf 0.799 \\
			& $\mu$TC (100\%)                  & 0.641 & 0.792 \\
			\midrule \multirow{2}{*}{German}
			& Mozeti{\v{c}} et al.~\cite{multilingual2016} & -     & 0.610 \\
			& B4MSA (89\%)                                 & 0.621 & 0.668 \\
			& $\mu$TC (89\%)                               & 0.614 & \bf 0.672 \\
			\midrule \multirow{3}{*}{Portuguese}
			& Mozeti{\v{c}} et al.~\cite{multilingual2016} & -     & 0.507 \\
			& B4MSA (58\%)                                 & 0.557 & 0.561 \\
			& $\mu$TC (58\%)                               & 0.562 & \bf 0.566 \\
			\midrule \multirow{3}{*}{Russian}
			& Mozeti{\v{c}} et al.~\cite{multilingual2016} & -     &  0.603 \\
			& B4MSA (69\%)                                 & 0.754 &  0.750 \\
			& $\mu$TC (69\%)                               & 0.754 &  \bf 0.751 \\
			\midrule \multirow{3}{*}{Swedish}
			& Mozeti{\v{c}} et al.~\cite{multilingual2016} & -   &  0.616 \\
			& B4MSA (93\%)                                 & 0.680  & \bf 0.691 \\
			& $\mu$TC (93\%)                               & 0.679  & 0.688 \\
			\midrule \multirow{2}{*}{Spanish}
			& B4MSA     & 0.657 & \bf 0.784 \\
			& $\mu$TC   & 0.649 & 0.780 \\
			\bottomrule
		\end{tabular}
	}
\end{table}

Finally, Table \ref{tab:spam} shows the results of spam classification task. Here, it can be seen that best results in the macro-F1 measure were obtained with our approach $\mu$TC; nevertheless, the best results in the accuracy score were achieved by Androutsopoulos et al. \cite{androutsopoulos2004learning} except in Ling-Spam dataset where $\mu$TC reached the best performance.

\begin{table}[ht]
	\centering
	\caption{spam classification}
	\label{tab:spam}
	\begin{minipage}{0.8\textwidth}
		\resizebox{\textwidth}{!}{
			\begin{tabular}{lcccc}
				\hline
				& \multicolumn{4}{c}{\textbf{macro-F1}} \\
				Data set & Androutsopoulos \cite{androutsopoulos2004learning} & Sakkis \cite{Sakkis2003} & Cheng \cite{HuaLi201288} & $\mu$TC\\  \cline{2-5}
				Ling-Spam 	& -  &  0.8957   	& 0.9870 &\textbf{0.9979}   \\
				PUA & 0.8897 & - & -			& \textbf{0.9478}\\
				PU1 & 0.9149 & - &\textbf{0.983}& 0.9664  \\
				PU2 & 0.6794 & - &	-			& \textbf{0.9044} \\
				PU3 & 0.9265 & - &\textbf{0.977}&0.9701 \\
			\end{tabular}
		}\end{minipage}
		\begin{minipage}{0.8\textwidth}
			\resizebox{\textwidth}{!}{
				\begin{tabular}{lcccc}
					\hline
					& \multicolumn{4}{c}{\textbf{accuracy}} \\
					Data set & Androutsopoulos \cite{androutsopoulos2004learningnuevo} & Sakkis \cite{Sakkis2003} & Cheng \cite{HuaLi201288}   & $\mu$TC          \\  \cline{2-5}
					Ling-Spam 	& -  			  &	- & 0.9800 & \textbf{0.9993}  \\
					PUA     	& \textbf{0.9600} & - & -      & 0.9482 \\
					PU1 		& \textbf{0.9750} & - &	0.971  & 0.9706 \\
					PU2   		& \textbf{0.9839} & - &	-	   & 0.9634 \\
					PU3  		& \textbf{0.9778} & - &	0.968  & 0.9738 \\
					\hline

				\end{tabular}
			}
		\end{minipage}
	\end{table}

	\subsection{About the pre-processing state of the input text}
	\label{sec:input-stage}
	\begin{table}
		\caption{The performance of $\mu$TC for text collections being in different stages of text normalization for \textsf{News} benchmark.}
		\label{tab:kfolds}
		\centering
		\resizebox{0.8\textwidth}{!}{
			\begin{tabular}{rcccc}
				\toprule
				kind of       &  actual   &  actual   &  pred     &  pred  \\
				preprocessing &  accuracy &  macro-F1 &  accuracy &  macro-F1 \\
				\midrule
				raw          & 0.8265 & 0.8199 &  0.8968 & 0.8963 \\
				all-terms    & 0.8340 & 0.8260 &  0.9075 & 0.9056 \\
				no-short     & 0.8310 & 0.8235 &  0.9052 & 0.9034 \\
				no-stopwords & 0.8373 & 0.8300 &  0.9099 & 0.9082 \\
				stemmed      & 0.8413 & 0.8344 &  0.9071 & 0.9058 \\
				\bottomrule
			\end{tabular}
		}
	\end{table}

	Here, the pre-processing step is analyzed; for this, Table \ref{tab:kfolds} shows different performances that correspond to the \textsf{News} benchmark in various stages of the normalization process, as used as inputs for $\mu$TC.
    We found that $\mu$TC achieves high performances without using additional sophisticated pre-processing steps, almost all of them, language dependent.
    For instance, using the raw text is just below $0.0148$ points than the performance using the {\em stemmed} collection. The human intervention to prepare the input text is barely needed by $\mu$TC without significantly reducing the performance in practice.
    Alternatively, methods like Escalante et al.~\cite{Escalante2015} and Cachopo~\cite{cardoso2007improving} need to use the stemmed version of the dataset to achieve its optimal performance, i.e., accuracy values ranging from $0.6623$ to $0.8460$, for more details see Table~\ref{tab:topic}.

	\subsection{On the robustness of the \textsf{score} function}
	\label{sec:cross-validation-study}

    The \textsf{score} function leads the model selection procedure to fulfill the requirements of the task. In this process, it is necessary to determine which precise quality's measure is needed, e.g., macro-F1 or accuracy. As any learning algorithm, it is necessary to protect the \textsf{score} with some validation schemes to avoid the latent overfitting. On this matter, we consider the use of two validation schemes: i) stratified $k$-folds and ii) a random binary partition of size $\beta n$ for the train set and $(1-\beta)n$ for the test set, for a (training) collection of size $n$.
    
    	\begin{figure}[!h]

		\subfigure[Authors NFL -- k-folds] {
			\includegraphics[width=0.5\textwidth]{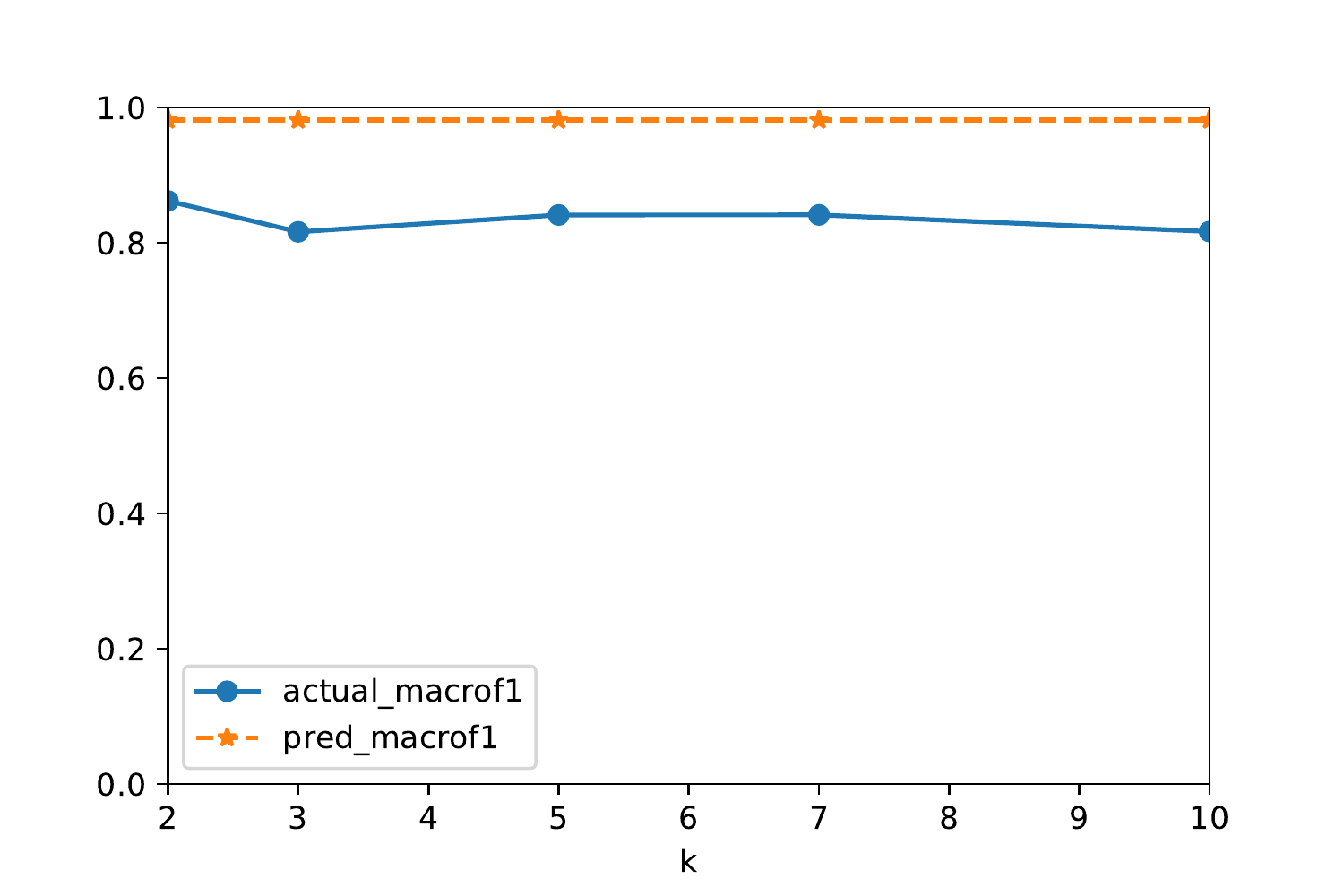}
			\label{fig:nfl-k}
		}\subfigure[Authors NFL -- binary partition] {
		\includegraphics[width=0.5\textwidth]{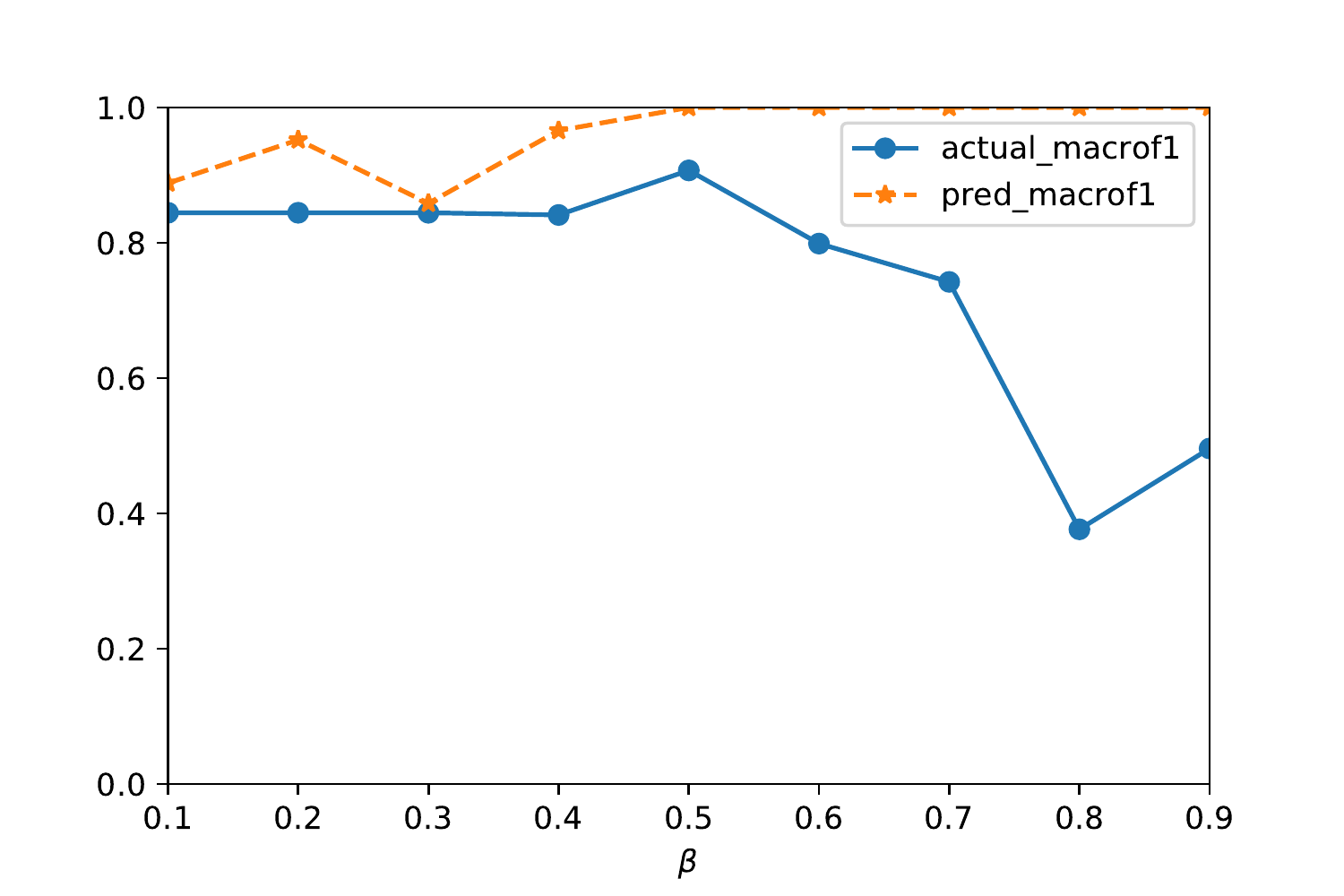}
		\label{fig:nfl-sampling}
	}

	\subfigure[Authors Business -- k-folds] {
		\includegraphics[width=0.5\textwidth]{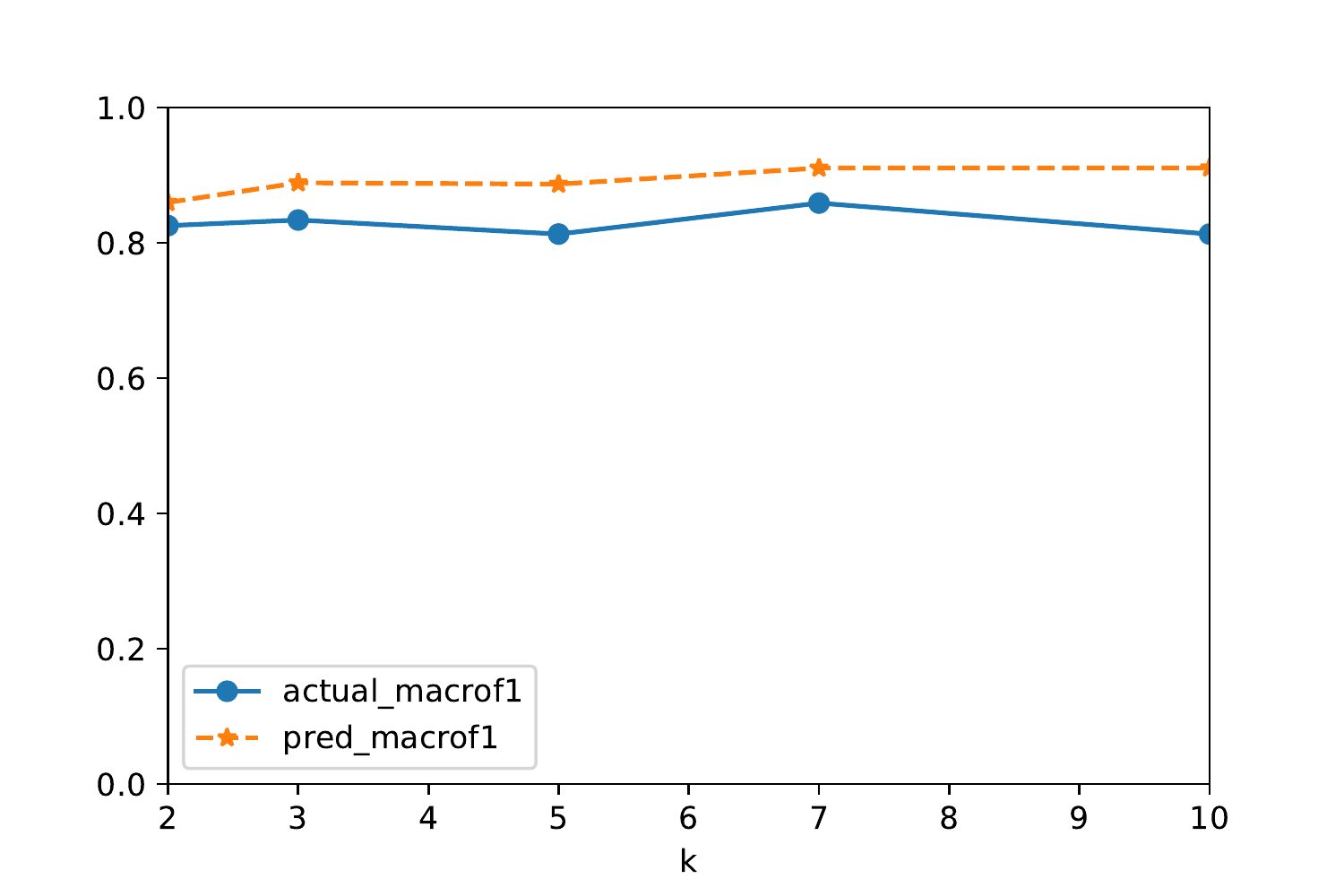}
		\label{fig:business-k}
	}\subfigure[Authors Business -- binary partition] {
	\includegraphics[width=0.5\textwidth]{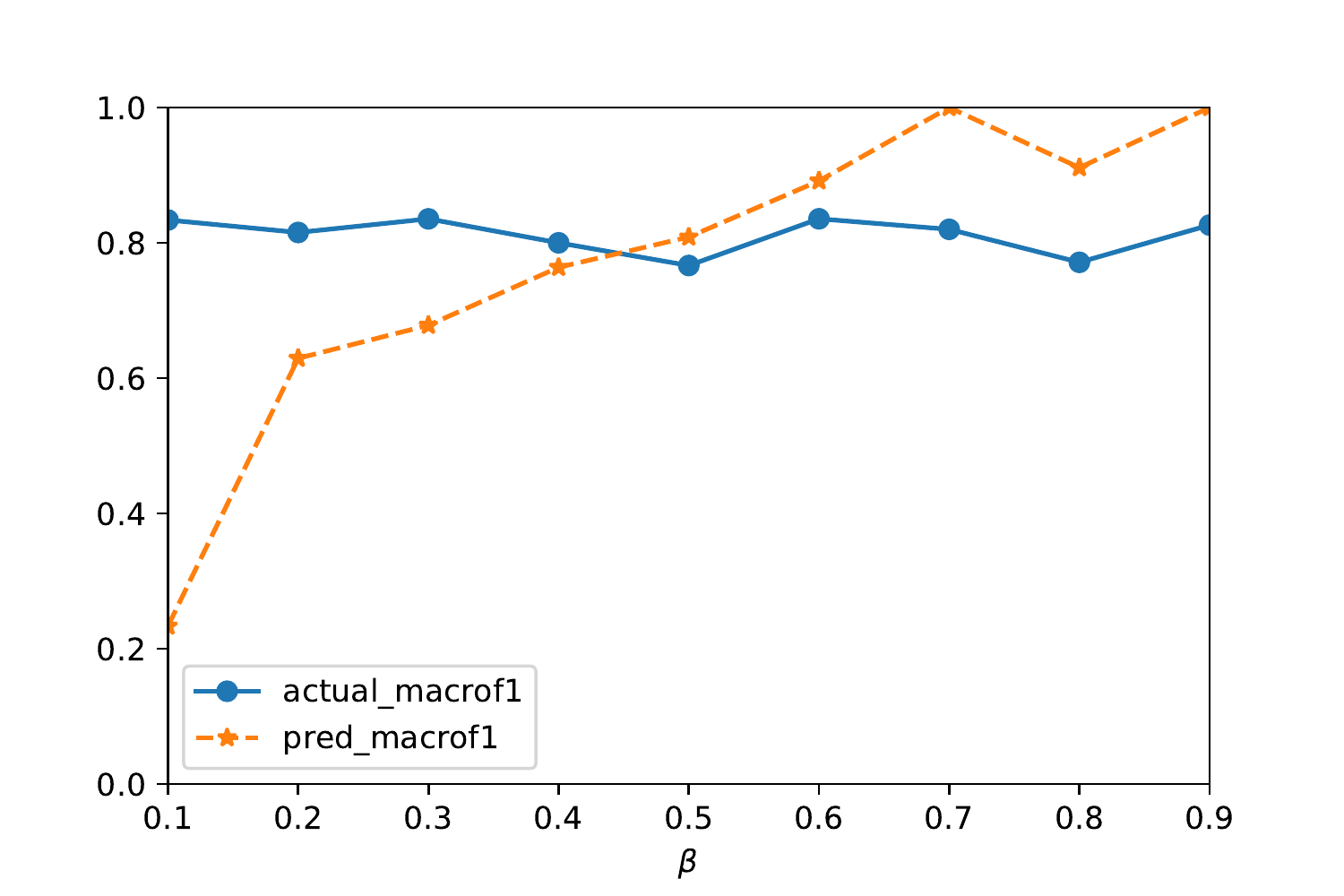}
	\label{fig:business-sampling}
}

\subfigure[Authors Cricket -- k-folds] {
	\includegraphics[width=0.5\textwidth]{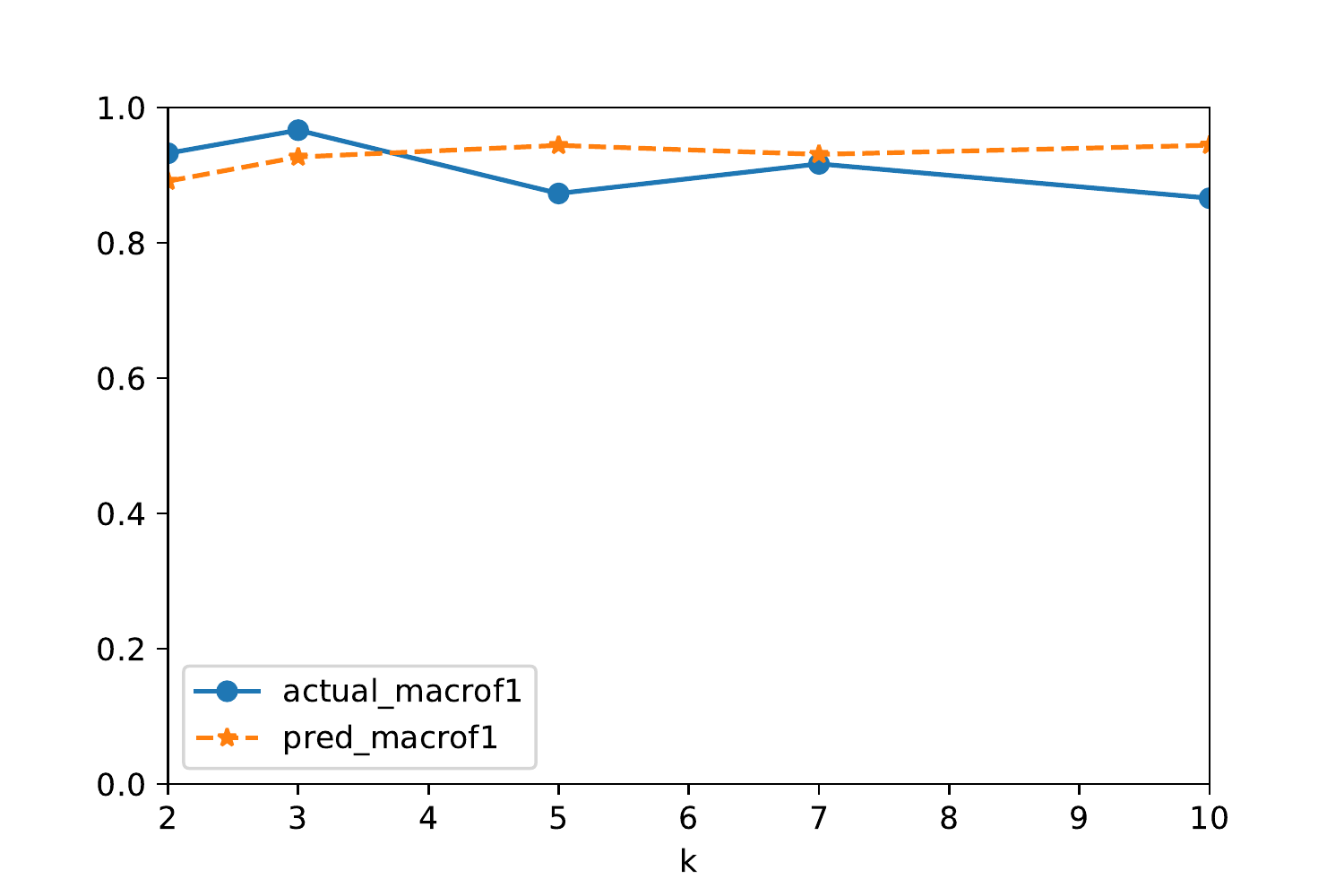}
	\label{fig:cricket-k}
}\subfigure[Authors Cricket -- binary partition] {
\includegraphics[width=0.5\textwidth]{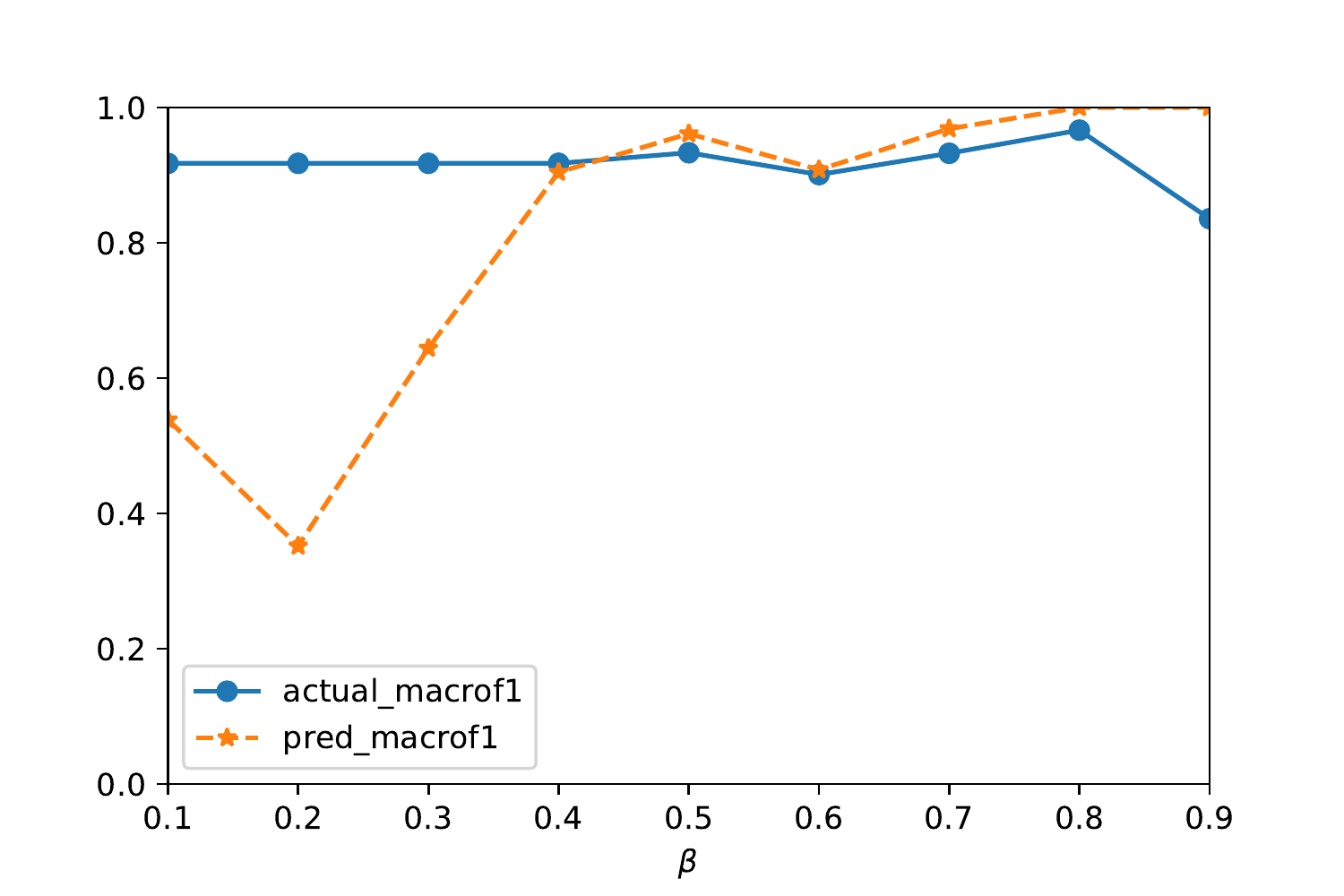}
\label{fig:cricket-sampling}
}
\caption{The final performance in small datasets as a function of the validation's stage of the \textsf{score} function of $\mu$TC; we consider two validation schemes for this purpose: i) $k$-folds and ii) random binary partitions of sizes $\beta n$ and $(1-\beta)n$, for training and testing subsets respectively.}
\label{fig:performance-small}
\end{figure}

\begin{figure}[!h]
	\subfigure[News -- k-folds] {
		\includegraphics[width=0.5\textwidth]{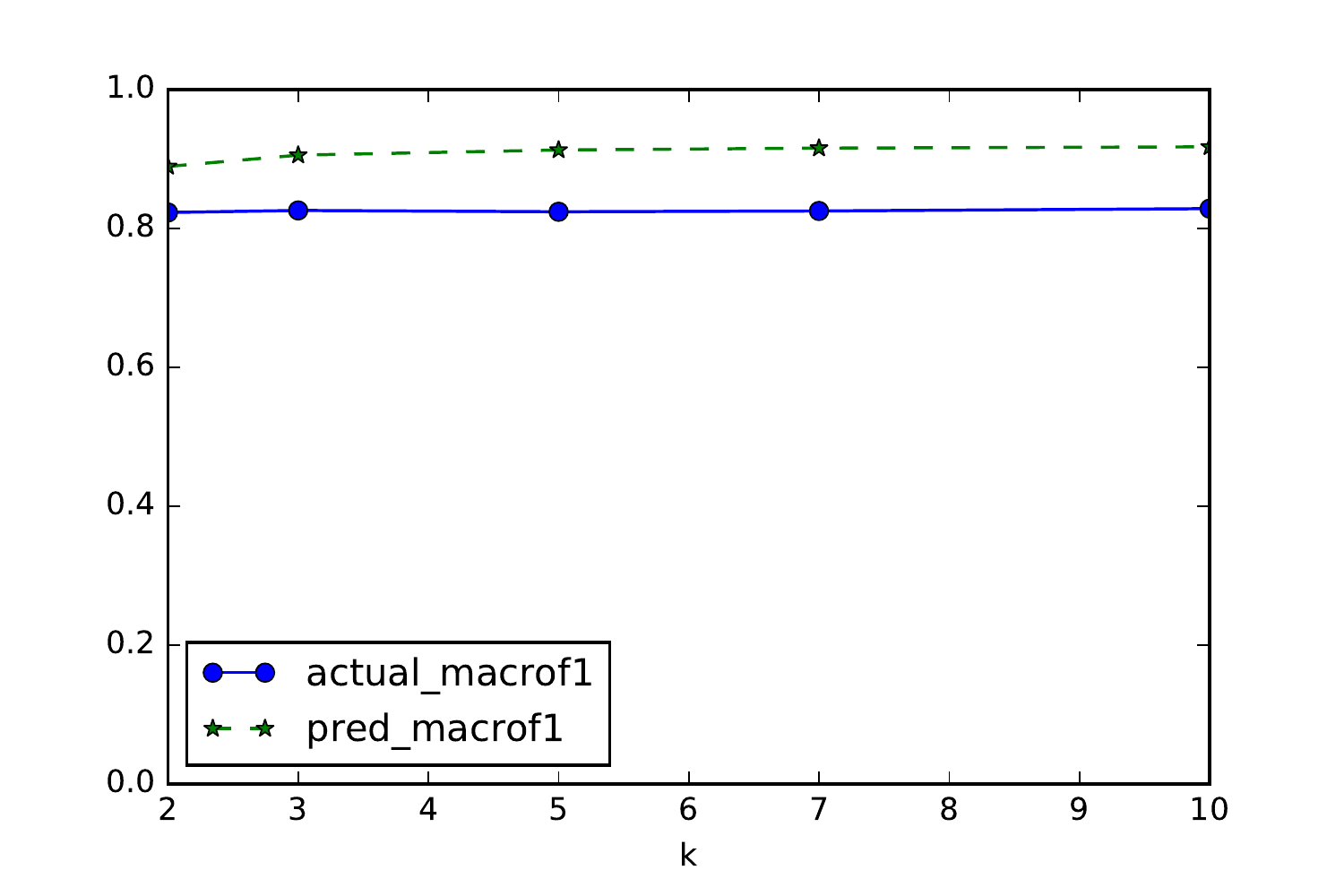}
		\label{fig:news-k}
	}~\subfigure[News -- binary partition] {
	\includegraphics[width=0.5\textwidth]{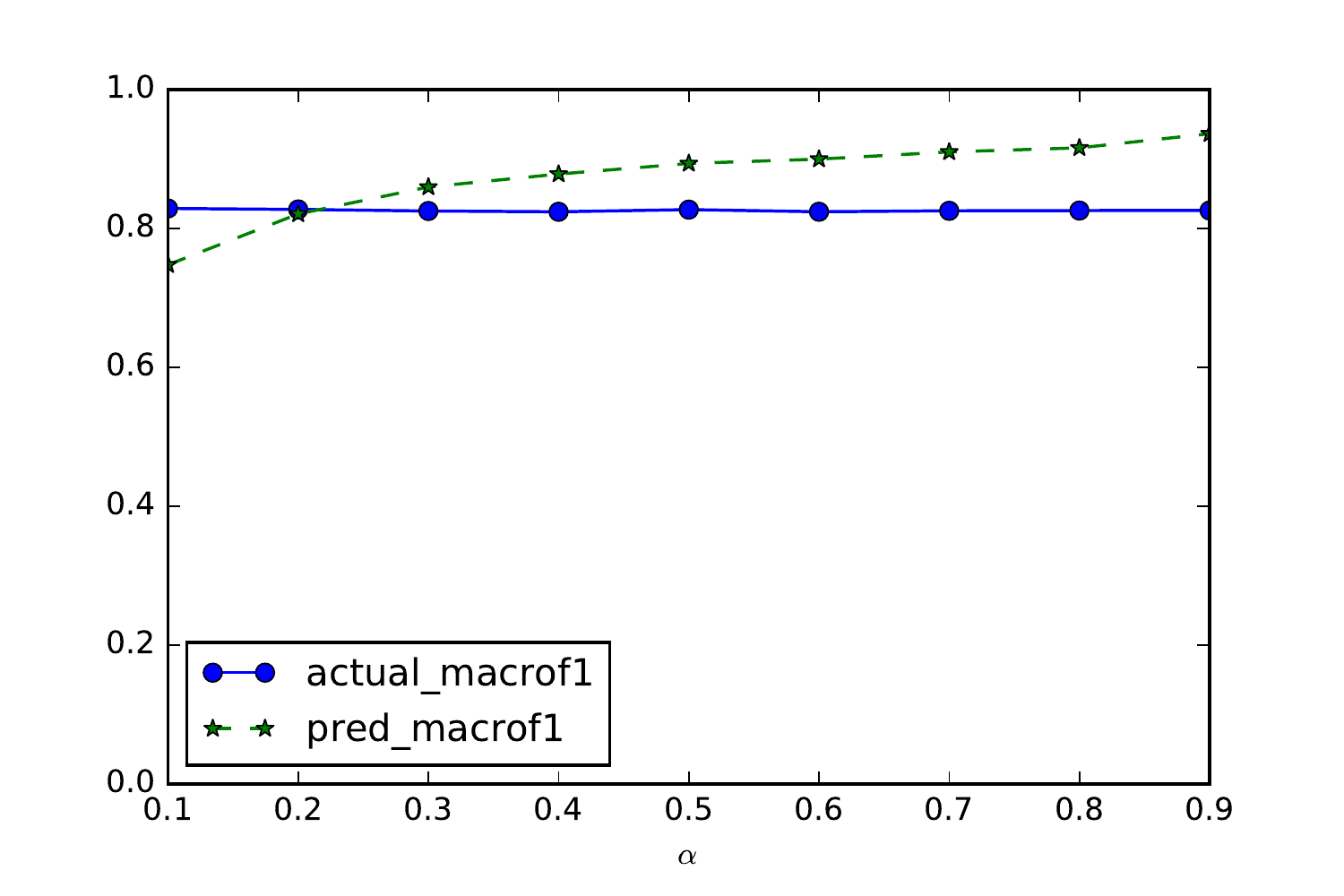}
	\label{fig:news-sampling}
}

\subfigure[WebKB -- k-folds] {
	\includegraphics[width=0.5\textwidth]{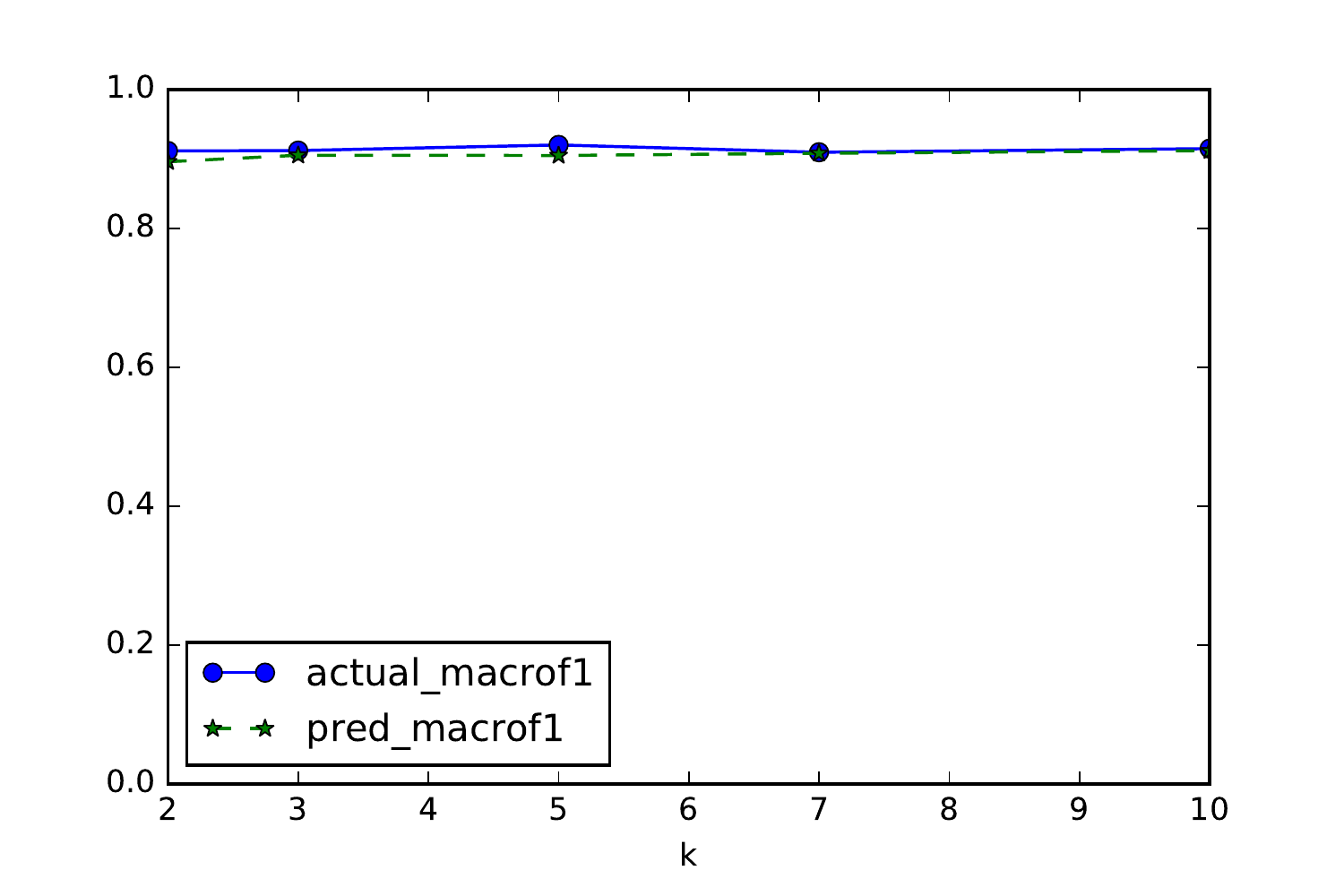}
	\label{fig:webkb-k}
}~\subfigure[WebKB -- binary partition] {
\includegraphics[width=0.5\textwidth]{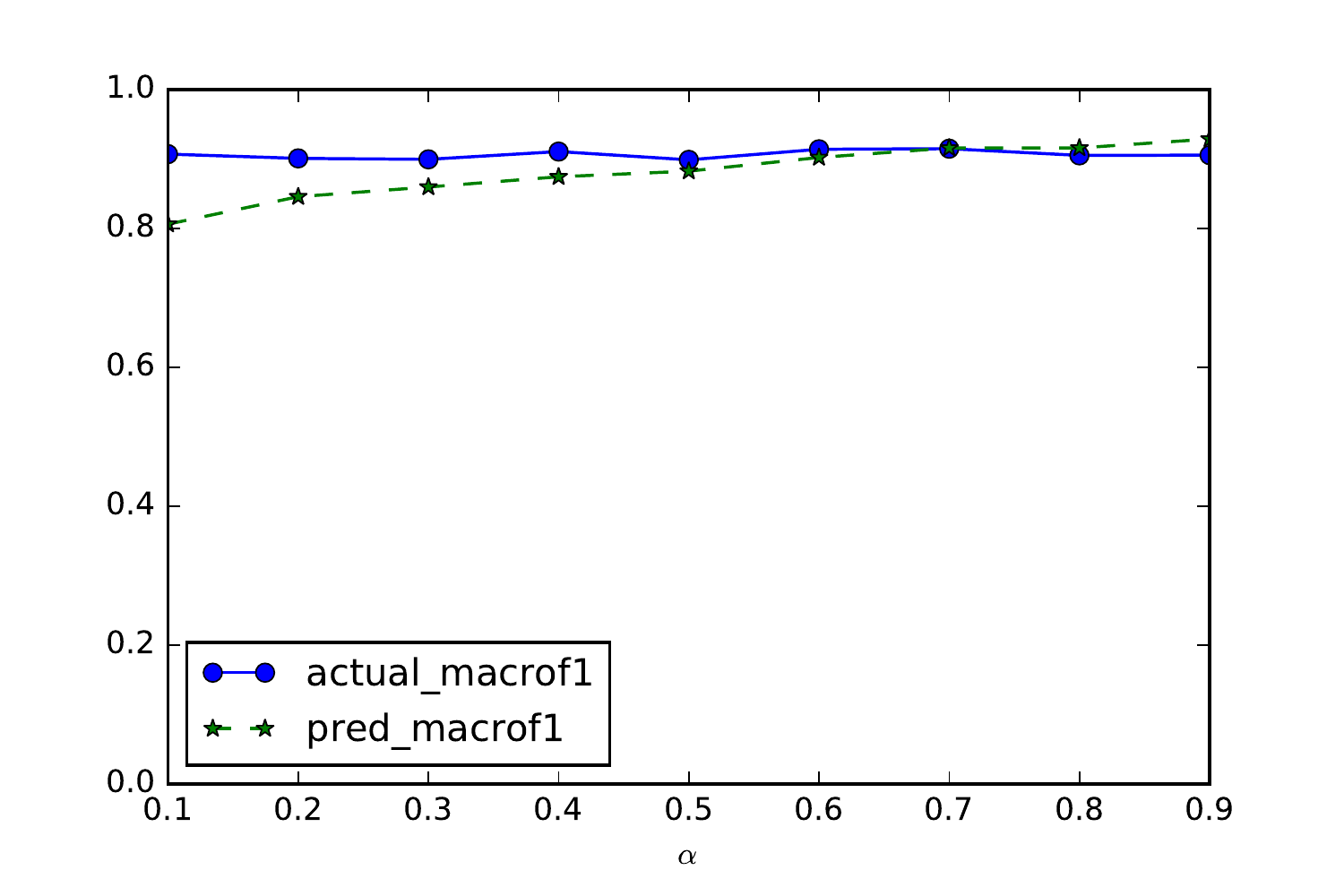}
\label{fig:webkb-sampling}
}

\subfigure[R52 -- k-folds] {
	\includegraphics[width=0.5\textwidth]{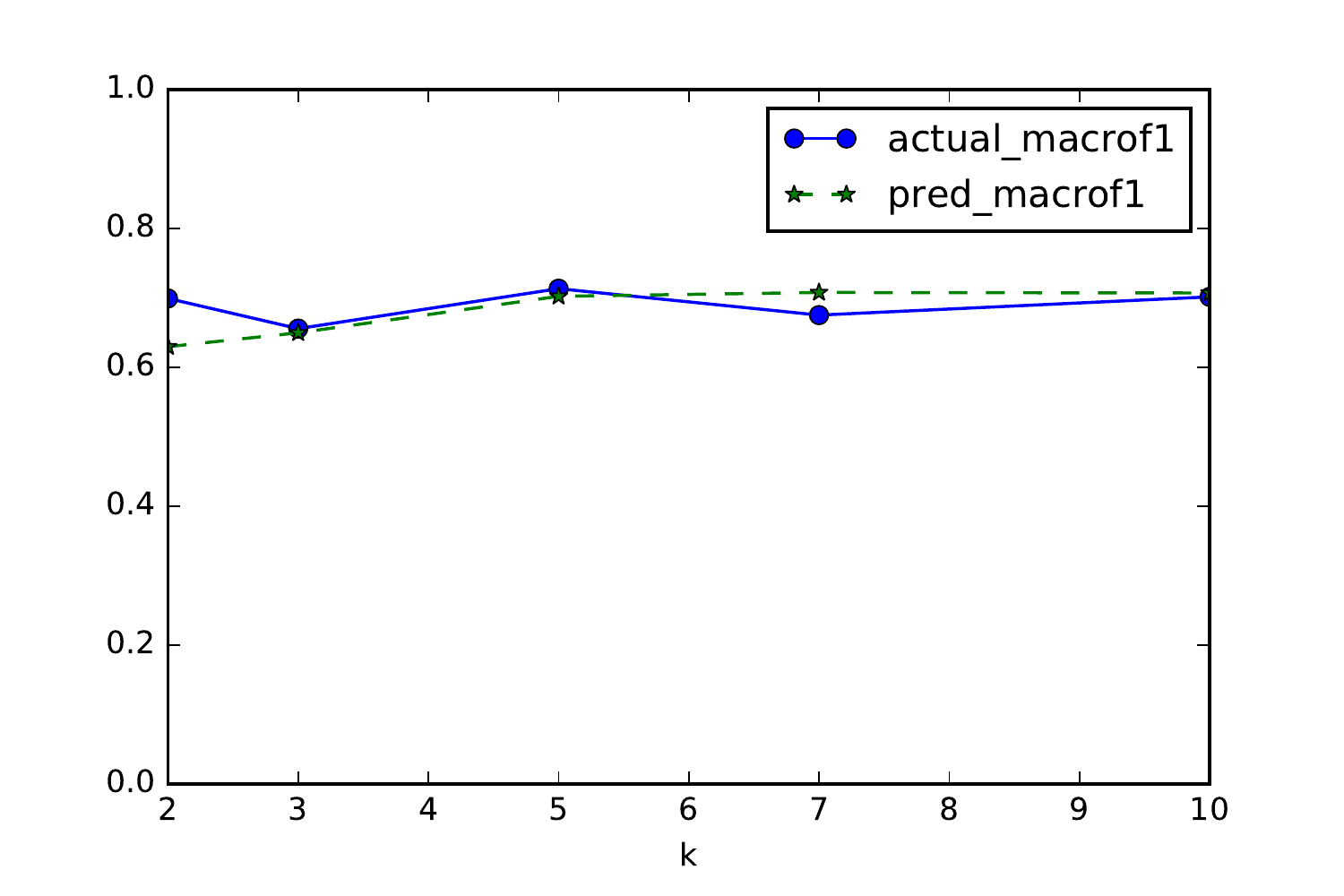}
	\label{fig:r52-k}
}\subfigure[R52 -- binary partition] {
\includegraphics[width=0.5\textwidth]{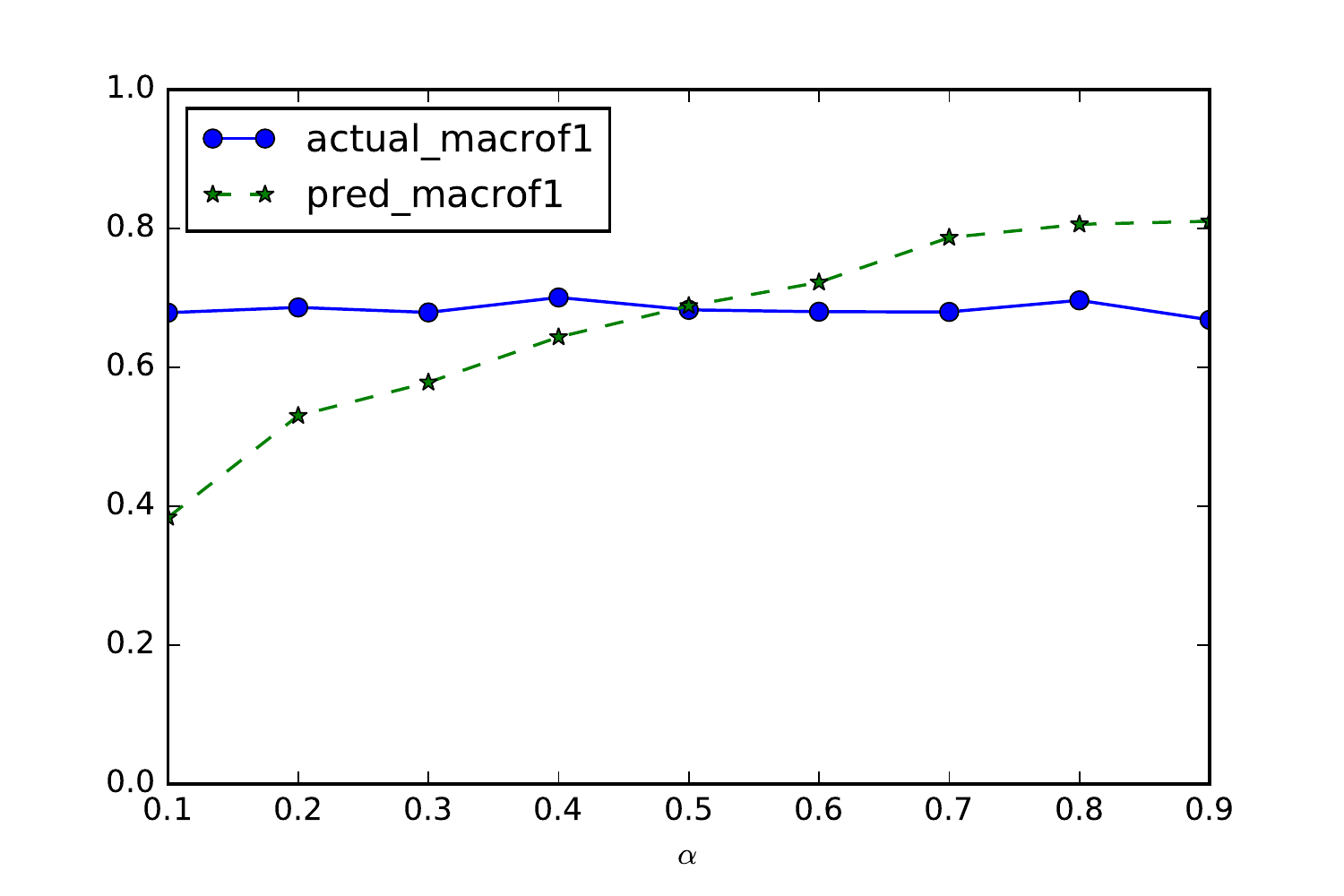}
\label{fig:r52-sampling}
}

\caption{The final performance on medium sized datasets as a function of the validation's stage of the \textsf{score} function of $\mu$TC; we consider two validation schemes for this purpose: i) $k$-folds and ii) random binary partitions of sizes $\beta n$ and $(1-\beta)n$, for training and testing subsets respectively.}
\label{fig:performance-medium}
\end{figure}

    To learn how to choose the right criteria, we review both the {\em predicted} and the {\em actual} performance (macro-F1, for instance) of these two validation schemes. The predicted macro-F1 is the performance achieved by the model selection procedure using some of the two mentioned validation schemes. The actual performance is the one obtained directly evaluating the gold-standard collection.

    Figure ~\ref{fig:performance-small} shows the performance of $\mu$TC on small databases. The stability of $k$-folds in terms of predicted and actual performance is supported by Figures~\ref{fig:nfl-k}, \ref{fig:business-k} and \ref{fig:cricket-k}. This is also true for larger datasets like those depicted in Figures~\ref{fig:news-k}, \ref{fig:webkb-k} and \ref{fig:r52-k}. The figures show that even on $k=2$ the $\mu$TC achieves almost its optimal actual performance; even when the predicted performance is most of the times better for larger $k$ values.
    On the other hand, the binary partition method is prone to overfit, especially on small datasets and small $1-\beta$ values (i.e., small test sets). For instance, Figure~\ref{fig:nfl-sampling} shows the performance for NFL; please note how $\beta = 0.5$ yields to very competitive performances, i.e. higher than 0.9 for both macro-F1 and accuracy. These performances are pretty higher than those achieved by the alternatives (see Table~\ref{tab:authorship}); however, $\beta > 0.5$ yields to low actual performances, contrasting the perfect predicted performance. A similar case happens for the Business dataset, Figure \ref{fig:business-sampling}; but in this case, the actual performance is relatively stable. The behaviour of binary partition in larger dataset is less prone to overfit, like Figures~\ref{fig:news-sampling} and \ref{fig:webkb-sampling} illustrate. Nonetheless, the case of R52, Figure~\ref{fig:r52-sampling}, shows that the overfitting issue is still latent; however, it barely affects the actual performance since the \textsf{score} function is applied to a large enough test set.

    As rule of thumb, it is safe to use $k$-fold cross-validation to compute \textsf{score} in the model selection stage. We encourage the use of small $k$ values (e.g., 2, 3 or 5) since the actual performance is relatively stable and the computational cost is kept low. Please notice that $k$-folds procedure introduces a factor of $k$ to the computational cost of \textsf{score}, and, algorithms to solve the underlying combinatorial optimization problem need to evaluate a considerable number of configurations to achieve good results.
    In cases where the number of samples is pretty large, or a rapid solution is required, the binary partition method is also a good choice, especially for high $1-\beta$ values. The later setup corresponds to prepare robust \textsf{score} functions at the cost of reducing the train set in the model selection stage. The reduction of the training set is not a major problem for the actual performance, as it is illustrated by experiments corresponding to binary partition performances, see Figures~\ref{fig:performance-small} and \ref{fig:performance-medium}.  Please remember, at this stage, we are just selecting a proper configuration, and in a subsequent step, the final model is computed using this configuration and the entire training dataset $\mathcal{D}$.


\section{Conclusions}
\label{sec:con}
In this work, a minimalistic and global approach to text classification is proposed. Moreover, our approach was evaluated in a broad range of classification tasks such as topic classification, sentiment analysis, spam detection and user profiling; for this matter, a total of $30$ databases related with these tasks were employed. In order to evaluate the performance of our approach, the results obtained in each task were compared to the state-of-the-art methods, related to each task. Additionally, we analyze the effect of the pre-processing stage. In this experiment, we observed that our approach is competitive with the alternative methods even using the raw text as input, without a penalty in the performance; therefore, it is possible to use $\mu$TC to create text classifiers with a little knowledge of natural language processing and machine learning techniques.
We also studied some simple strategies to avoid overfitting problem; we consider using a $k$-fold cross-validation scheme and a binary partition to perform the model selection. Based on our experimental observation, our $\mu$TC can both properly fit the dataset and speedup the construction step using small $k$ values in cross-validation schemes and small training sets when we use binary random partitions. We also found that perform $k$-folds can be the preferred validation scheme on small to medium sized datasets, but very large datasets can use the binary partition scheme without a significant reduction of the performance, and also, keeping the cost the entire process low.


\bibliographystyle{plain}

\end{document}